%% file: main.tex
\documentclass[11pt]{article}
\usepackage[margin=1in]{geometry}
\usepackage[T1]{fontenc}
\usepackage{mathptmx}
\usepackage[scaled=.92]{helvet}
\usepackage{courier}
\usepackage{amsmath,amssymb}
\usepackage{booktabs}
\usepackage{float}
\usepackage{graphicx}
\usepackage{xcolor}
\usepackage[skins]{tcolorbox}
\usepackage{caption}
\usepackage{enumitem}
\usepackage{algorithm}
\usepackage{algpseudocode}
\usepackage{tikz}
\usetikzlibrary{arrows.meta,positioning,calc,fit}
\usepackage[numbers,sort&compress]{natbib}
\usepackage{microtype}
\usepackage{tabularx}
\usepackage[colorlinks=true,linkcolor=blue!50!black,citecolor=blue!50!black,urlcolor=blue!50!black]{hyperref}

\setlist{itemsep=1pt,topsep=3pt}
\newcommand{\tn}[1]{\texttt{#1}}
\newcommand{\SiteURL}{https://joessattes.github.io/ai-model-explorer/}
\newcommand{\CodeURL}{https://github.com/JoesSattes/ai-model-explorer}
\newcommand{\PypiURL}{https://pypi.org/project/ai-model-explorer/}
\newcommand{\DataURL}{https://huggingface.co/datasets/JoesSattes/ai-vocabulary-tax}
\input{numbers}

\input{numbers_base}
\input{numbers_extra}

\input{numbers_quality}

\begin{document}
\thispagestyle{plain}
\begin{tcolorbox}[enhanced,colback=black!4,colframe=black!4,arc=10pt,boxrule=0pt,left=16pt,right=16pt,top=14pt,bottom=12pt]
{\sffamily\bfseries\LARGE The Vocabulary Tax:\\[2pt] Safely Trimming Unused Tokens from LLMs\\[2pt] for On-Device Use\par}
\vspace{10pt}
{\sffamily\bfseries\small Sattaya Singkul}\par
\vspace{2pt}
{\small Independent Researcher, Bangkok, Thailand \quad \href{mailto:sattaya.singk@gmail.com}{\texttt{sattaya.singk@gmail.com}}\par}
\vspace{10pt}
{\small\setlength{\parskip}{0pt}\noindent Open large language models (LLMs) and speech models ship vocabularies of up to 262k tokens, and every user downloads the tables those tokens index (the input embedding, an untied output head and, in recent on-device models, per-layer embeddings), whatever languages they need. We call the bytes in rows a deployment never uses the \emph{vocabulary tax} and measure it, without downloading weights, for \NGen{} generative and speech-recognition models on the Hugging Face Hub. For a Thai and English deployment it is a median \TaxSA{} of model bytes below 0.5B parameters and \TaxSD{} above 8B, and \ExGemmaFourSGB{}\,GB in Gemma~4 E2B: it falls hardest on the small models that edge devices run. Keeping only the rows of the deployment's writing systems is exact. Keeping the tokens seen in the user's own text removes much more, a median \TaxCA{} below 0.5B, but the usual way of doing it breaks BPE tokenizers: it drops the intermediate tokens that merges build on, so new text is split differently \NaiveRatio{} times as often as the standard check reports. We repair this with merge closure, at a cost of \ExtraRows{} more rows, and add a \emph{trim certificate}: a Good--Turing estimate, from the trimming text alone, of how much new text the trimmed tokenizer will split differently (correlation \CertR{} over \CertN{} settings in 22 languages) and of how much more text a target needs. Without test data our method never exceeds a 2\% target, while prior keep rules checked on test data exceed it in \ViolFreqtwo--\ViolVTtwo{} of 44 settings. On six models of 270M to 3B parameters in 11 languages, trimmed models keep their predictions: where the certificate passes, a median \QOkRm{} of rows is removed and bits per character rise by at most 2.0\%. The census, the trimming and the certificate are released as an open Python package, and the per-model tax for 41 languages as an open dataset.\par}
\vspace{10pt}
{\footnotesize\sffamily
\begin{minipage}[b]{0.76\linewidth}
\textbf{Date:} \rmfamily November 2026\\
\sffamily\textbf{Playground:} \href{\SiteURL}{\ttfamily\SiteURL}\\
\textbf{Code:} \href{\CodeURL}{\ttfamily\CodeURL}\\
\textbf{Data:} \href{\DataURL}{\ttfamily huggingface.co/datasets/JoesSattes/ai-vocabulary-tax}\\
\textbf{Python:} \href{\PypiURL}{\ttfamily pip install ai-model-explorer}
\end{minipage}\hfill
\begin{minipage}[b]{0.22\linewidth}\raggedleft
\includegraphics[height=0.9cm]{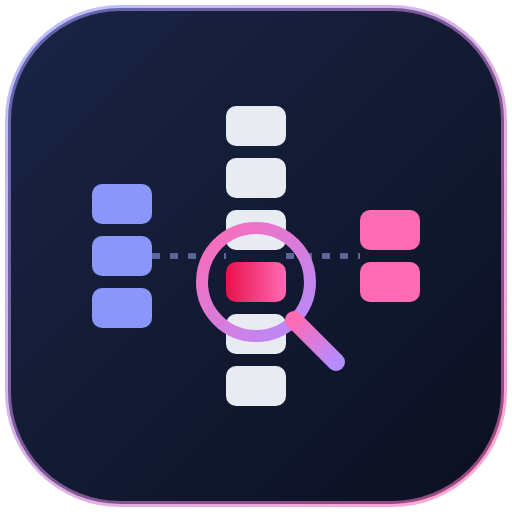}\\[2pt]
{\sffamily\bfseries\small AI Model Explorer}
\end{minipage}}
\end{tcolorbox}
\vspace{6pt}

\section{Introduction}

A team that deploys an open model to a phone, a call-center server or a kiosk usually serves a handful of languages. The model it downloads was trained for many more. Its vocabulary covers dozens of writing systems, and every table indexed by the vocabulary has one row per piece: the input embedding, the output head when it is not tied to the embedding, and, in recent on-device models such as Gemma~3n and Gemma~4, per-layer embedding tables that hold nearly half of the model's parameters. A Thai and English deployment of Gemma~4 E2B stores \ExGemmaFourSGB{}\,GB of rows that no Thai or English text can ever select. We call these bytes the \emph{vocabulary tax}.

The tax is easy to state and has not been measured. Work on tokenizer fairness measures how many tokens a sentence costs in each language~\citep{petrov2023unfairness,ahia2023cost,rust2021tokenizer}, a cost in compute and context length. Work on vocabulary trimming removes unused rows from particular models and shows that accuracy on the kept languages is preserved~\citep{ushio2023vt,bogoychev2024updowns,squeezebits2025trim,gurgurov2025compression}. Neither says how large the memory cost is across the models people download, who pays it, or how a user can tell, before deploying, whether a trimmed model will receive their text as the original did.

We answer these questions by building on our census of 12,974 Hugging Face repositories, which reads each model's anatomy from its weight headers alone~\citep{singkul2026manifest}. In doing so we find that the standard corpus-based way of trimming is unsound for the tokenizers most open models now use. Trimming by corpus keeps the vocabulary pieces that appear as tokens in a sample of target text. In byte-pair encoding (BPE)~\citep{sennrich2016bpe}, a piece is produced only by merging two smaller pieces, and many of these smaller pieces never appear as tokens because they are always merged further. A corpus rule removes them, and every piece built on them becomes unreachable: its row is kept but can never be selected, and the words it covered fall apart into smaller pieces than the trimming intended. The damage is invisible to the usual check, which counts the tokens of held-out text whose rows were removed.

The contributions are as follows.
\begin{enumerate}[leftmargin=*]
\item \textbf{A weight-free census of the vocabulary tax} for \NGen{} generative and speech-recognition models (Section~\ref{sec:rq1}). The tax is regressive: the median falls from \TaxSA{} of model bytes below 0.5B parameters to \TaxSD{} above 8B ($\rho=-0.34$). Across tokenizer families it trades against the compute tax of long token sequences.
\item \textbf{An exact keep rule by writing system} and the conditions that make it exact, including Latin pieces, without which 3.5--6.6\% of Thai tokens are lost (Section~\ref{sec:rq2}).
\item \textbf{Merge closure} (Section~\ref{sec:closure}). Across 22 languages and four BPE families, corpus trimming leaves a median of \OrphMed{} kept rows (\OrphShare) unreachable, and the real share of new text that is split differently is a median \NaiveRatio{} times the nominal share, in \NaiveWorse{} of \NaiveMeas{} measurable settings more than 1.2 times; sequences grow by up to \NaiveTokMax. Adding the merge parents of every kept piece costs a median \ExtraRows{} more rows and brings the real rate to \ClosedRatio{} of the nominal one. Against four keep rules of prior work, prior rules re-split \MatchVT--\MatchFreq{} times as much new text at the same compression, and ours removes the most rows at a 1\% target (Section~\ref{sec:baselines}).
\item \textbf{A trim certificate} (Section~\ref{sec:cert}): a Good--Turing estimate of the real re-split rate of new text and an extrapolation of how much text would bring it under a target, both from the trimming text alone, validated against rebuilt tokenizers; its error is 4.5--6.6\%, against 23--27\% for hold-out and cross-validation estimates.
\item \textbf{A 22-language study} (Section~\ref{sec:rq3}): every language can remove about the same share of bytes, but the text needed to do so safely differs by two orders of magnitude, ordered by how much of the language's vocabulary it shares with English. The certificate also holds per language in mixed-language deployments (median ratio \MixPer; Section~\ref{sec:mix}), and we quantify how it degrades under domain shift and how to guard against it.
\item \textbf{A quality evaluation of trimmed models} (Section~\ref{sec:quality}): six edge-sized models, 11 languages, bits per character, greedy continuations and Belebele. The script rule leaves the models' predictions unchanged up to renormalization; under the corpus rule, the loss in bits per character is about 2.5 times the re-split rate, so the certificate, computed before any model is run, also predicts the quality cost.
\end{enumerate}
We also report how many edge-sized models fall under common memory budgets once trimmed (Section~\ref{sec:rq5}). The measurement, the merge-closed trimming and the certificate are part of the open Python package \tn{ai-model-explorer}, and the corpus rule with its certificate also runs in the browser on the user's own text, without uploading it. The tax of 2,061 Hub models for 41 languages, the tokenizer-family results, the certificate and merge-closure measurements and the quality evaluation are released as a dataset (\url{\DataURL}).

\section{Related work}
\label{sec:related}

\paragraph{Vocabulary trimming.} \citet{ushio2023vt} remove the vocabulary entries of a multilingual model that do not appear in a target-language corpus and show that accuracy on the target language is retained without retraining; their models use Unigram vocabularies, which have no merges. \citet{bogoychev2024updowns} compare a script-based and a corpus-based selection of sub-vocabularies at inference time, the two rules we study, and measure memory and speed gains for several models. SqueezeBits~\citep{squeezebits2025trim} report latency gains of 15--21\% for Gemma~3 1B. \citet{gurgurov2025compression} keep the most frequent pieces of a target-language corpus as part of compressing multilingual encoders, COMPACT~\citep{kwek2025compact} removes rare tokens together with the merge rules that produce them, alongside channel pruning, VocabTailor~\citep{zhang2025vocabtailor} selects an output vocabulary per task and input, and \citet{benshoham2026spec} trim the vocabulary of a speculative-decoding draft model. These works establish, on the models they chose, that trimming preserves accuracy when the kept set covers the target text. None examines whether a frequency-selected set of BPE pieces can still be produced by the rebuilt tokenizer, and none tells a user, without held-out data, whether their text is enough. Under-trained tokens are related: \citet{land2024magikarp} find that intermediate BPE pieces are often under-trained because they rarely occur as tokens, the same property that makes a corpus rule remove them.

\paragraph{Tokenization cost across languages.} \citet{petrov2023unfairness} and \citet{ahia2023cost} show that the same content costs up to an order of magnitude more tokens in some languages than in English, and \citet{rust2021tokenizer} relate the quality of a multilingual tokenizer for a language to downstream performance. This is the compute side of the cost of a multilingual vocabulary. We measure the memory side and show that the two trade against each other across tokenizer families (Figure~\ref{fig:twotax}).

\paragraph{Estimating unseen mass.} The Good--Turing estimator~\citep{good1953,gale1995gt} estimates the probability of events not seen in a sample from the number of events seen once. Heaps' law~\citep{heaps1978} describes how the number of distinct types grows with sample size. We apply both to a question they have not been used for: how much of a deployment's future text will fall outside a vocabulary chosen from a sample of that text.

\paragraph{Reading models without weights.} The tensor manifest of a safetensors file~\citep{safetensors} records the name, shape and storage type of every tensor. Our previous work recovers a model's anatomy from the manifest alone and reports a census of 12,974 Hub repositories~\citep{singkul2026manifest}; we take the token tables and model sizes of this paper from that census.

\section{Problem and method}
\label{sec:method}

\subsection{The vocabulary tax}
A model $\mathcal{M}$ stores token tables $E_1,\dots,E_k$, each with $V$ rows (the vocabulary size) and $b_j$ bytes. They are found in the manifest as the tensors whose leading dimension equals $V$ and that are used as an input embedding, an untied output head, or a per-layer embedding~\citep{singkul2026manifest}. A deployment for a set of languages $L$ needs a set $K\subseteq\{1,\dots,V\}$ of rows. The vocabulary tax of $\mathcal{M}$ for $L$ is
\begin{equation}
\mathrm{tax}(\mathcal{M},L)=\frac{1}{B(\mathcal{M})}\sum_{j=1}^{k} b_j\,\frac{|R|}{V},\qquad R=\{1,\dots,V_{\mathrm{tok}}\}\setminus K,
\end{equation}
where $B(\mathcal{M})$ is the stored size of the model and $V_{\mathrm{tok}}\le V$ is the number of pieces in the tokenizer file. Rows past the end of the tokenizer are kept: they may be padding or special tokens that the file does not list. The tax depends on the keep rule that defines $K$.

\subsection{Two keep rules}
\paragraph{Script rule.} Each piece is decoded to text (undoing byte-level encoding and the SentencePiece word marker). A piece is kept if every character in it has a Unicode script~\citep{unicode_uax24} in the scripts of $L$, or in the \emph{Common} or \emph{Inherited} scripts, or is a format character (category Cf), and pieces that are special tokens, byte fallbacks, digits or punctuation are always kept. Latin is added to every $L$ by default because names, numbers, code and loanwords appear in Latin letters in the text of every language. Under this rule, any text written in the scripts of $L$ is tokenized into kept pieces only, so it is tokenized exactly as before, provided the tokenizer is rebuilt consistently (below). Each of the inclusions repairs a failure we observed: the Japanese length mark \textit{ch\=oon} (U+30FC) and the Arabic \textit{tatweel} (U+0640) belong to the Common script, combining marks to Inherited, and zero-width spaces to Cf.

\paragraph{Corpus rule.} Given lines of target text $C$, keep a piece if the script rule keeps it and, in addition, it is a special token, it is a single character, or it occurs at least $m$ times in the tokenization of $C$ (we use $m=1$ unless stated). The single-character pieces guarantee that any word can still be spelled, so no text fails to encode. Pieces seen fewer than $m$ times are removed, and any later word that needed them is split into smaller kept pieces, which changes the model's input. We call the share of tokens in new text that fall in removed rows the \emph{re-split rate}; it is zero for the script rule on text in $L$'s scripts.

\subsection{Rebuilding the tokenizer and merge closure}
\label{sec:closuremethod}
Trimming rows is only safe if the rebuilt tokenizer maps text to the intended pieces. For BPE tokenizers~\citep{sennrich2016bpe,radford2019gpt2} we remove the vocabulary entries of removed pieces and every merge whose inputs or output were removed; for SentencePiece models~\citep{kudo2018sentencepiece} we remove the pieces from the model protocol buffer. Kept pieces are re-indexed in their original order, and a list of kept ids maps new ids to original rows, so the token tables are sliced with one indexing operation.

A BPE piece $p$ is created by one merge $p=a\,b$ of two pieces, its \emph{merge parents}, and the tokenizer can produce $p$ only if it can produce $a$ and $b$. A keep set $K$ is \emph{merge-closed} if it contains the parents of each of its pieces. The script rule is closed by construction for whole characters, since the parents of a piece are substrings of it and have the same scripts; the corpus rule is not, because it keeps pieces by their frequency \emph{as tokens}, and the parents of a frequent piece are often never tokens themselves. In a set that is not closed, a kept piece whose parent was removed is an \emph{orphan}: its row is never selected, and every word that used it is split into smaller pieces. We therefore close every keep set: starting from $K$, we add the parents of each kept piece, recursively. The parents are read from the merge list of \tn{tokenizer.json} and GGUF files, and for tiktoken and Tekken vocabularies, which store ranks instead of merges, from the last merge that BPE performs when it encodes the piece by itself.

\subsection{The trim certificate}
\label{sec:certmethod}
Under the corpus rule, the user needs to know the \emph{real} re-split rate of text they have not yet seen: the share of the tokens the original tokenizer would produce that the trimmed tokenizer does not reproduce. Holding out part of $C$ measures it, but costs text and, as we show, overestimates the rate of the final keep set, which is built from all of $C$. Instead we estimate it from $C$ itself. Let $N$ be the number of tokens in the tokenization of $C$, and $N_c$ the number of distinct candidate pieces (kept by the script rule, not special and longer than one character) that occur exactly $c$ times. The corpus rule removes the candidates seen fewer than $m$ times. By the Good--Turing estimate~\citep{good1953}, the probability that a new token is a candidate seen $c$ times in $C$ is $(c+1)N_{c+1}/N$, so the re-split rate of new text from the same source is estimated by
\begin{equation}
\hat r(C)=\sum_{c=0}^{m-1}\frac{(c+1)\,N_{c+1}}{N},\qquad \hat r=\frac{N_1}{N}\ \text{for}\ m=1.
\label{eq:gt}
\end{equation}
To say how much text would reach a target rate $\tau$, we compute $\hat r$ on the first half and on all of $C$ ($n$ lines), fit $\hat r(n)\propto n^{-\beta}$, which is the form Heaps' law~\citep{heaps1978} gives for a Zipfian sample, and report
\begin{equation}
n^{*}=n\left(\hat r(n)/\tau\right)^{1/\beta},\qquad \beta=\frac{\log\big(\hat r(n/2)/\hat r(n)\big)}{\log 2}.
\label{eq:need}
\end{equation}
The certificate reports $\hat r$, whether it is below $\tau$ (default 1\%), and $n^{*}$ when it is not. It is computed in the time the tokenization of $C$ takes. Algorithm~\ref{alg:cert} summarizes the corpus rule with merge closure and the certificate.

\begin{algorithm}[t]
\caption{Merge-closed corpus trimming with a certificate}
\label{alg:cert}
\small
\begin{algorithmic}[1]
\Require tokenizer $T$ with pieces $1..V_{\mathrm{tok}}$, languages $L$, text $C$ ($n$ lines), minimum count $m$, target $\tau$
\State $K_s \gets$ script rule$(T, L\cup\{\text{Latin}\})$ \Comment{specials, bytes, Common, Inherited, Cf always kept}
\State $\mathrm{cnt} \gets$ piece counts of $T(C)$;\quad $N \gets \sum \mathrm{cnt}$
\State $K_c \gets \{p\in K_s : \mathrm{special}(p) \lor |p|\le 1 \lor \mathrm{cnt}(p)\ge m\}$
\State \textbf{repeat} add the merge parents of every $p\in K_c$ to $K_c$ \textbf{until} no change \Comment{merge closure}
\State $\hat r(n) \gets \sum_{c<m}(c+1)N_{c+1}/N$ over candidates of $K_s$ \Comment{Eq.~\ref{eq:gt}}
\State $\hat r(n/2) \gets$ the same on the first $n/2$ lines;\quad $\beta \gets \log_2(\hat r(n/2)/\hat r(n))$
\State $n^{*} \gets n\,(\hat r(n)/\tau)^{1/\beta}$ if $\hat r(n)>\tau$ \Comment{Eq.~\ref{eq:need}}
\State rebuild $T$ on $K_c$ (drop merges touching removed pieces); kept ids $\gets K_c$ in original order
\State verify: test sentences of $L$ tokenize identically under the script rule
\State \Return $T'$, kept ids, $\hat r(n)$, $[\hat r(n)\le\tau]$, $n^{*}$
\end{algorithmic}
\end{algorithm}

\begin{figure}[t]
\centering
\resizebox{0.78\linewidth}{!}{\begin{tikzpicture}[font=\sffamily\scriptsize, node distance=4mm and 7mm,
  box/.style={draw=black!35, rounded corners=3pt, fill=black!3, align=center, inner sep=4pt, minimum height=9mm},
  hl/.style={draw=blue!50!black, rounded corners=3pt, fill=blue!6, align=center, inner sep=4pt, minimum height=9mm},
  arr/.style={-{Stealth[length=2mm]}, black!55}]
\node[box] (man) {Tensor manifest\\(header, KB)};
\node[box, below=of man] (tok) {Tokenizer file\\(any format)};
\node[box, right=of man] (tab) {Token tables\\embed, head, PLE};
\node[box, right=of tok] (ks) {Script rule\\$K_s$ (exact)};
\node[box, below=of ks] (txt) {User text $C$};
\node[box, right=of ks] (kc) {Corpus rule\\+ merge closure};
\node[hl, right=of tab] (tax) {Vocabulary tax\\bytes, share};
\node[hl, right=of kc] (cert) {Certificate\\$\hat r$, lines needed};
\node[box, below=of cert] (trim) {Rebuilt tokenizer,\\kept ids, slicing};
\draw[arr] (man) -- (tab); \draw[arr] (tok) -- (ks); \draw[arr] (ks) -- (kc); \draw[arr] (txt) -| (kc);
\draw[arr] (tab) -- (tax); \draw[arr] (ks) -- (tax); \draw[arr] (kc) -- (tax.south);
\draw[arr] (kc) -- (cert); \draw[arr] (kc) |- (trim);
\end{tikzpicture}}
\caption{Pipeline. The tax is computed from the manifest and the tokenizer without weights. The corpus rule refines the script rule with the user's text, and the certificate estimates, from that text alone, how much new text the trimmed tokenizer will split differently.}
\label{fig:pipeline}
\end{figure}
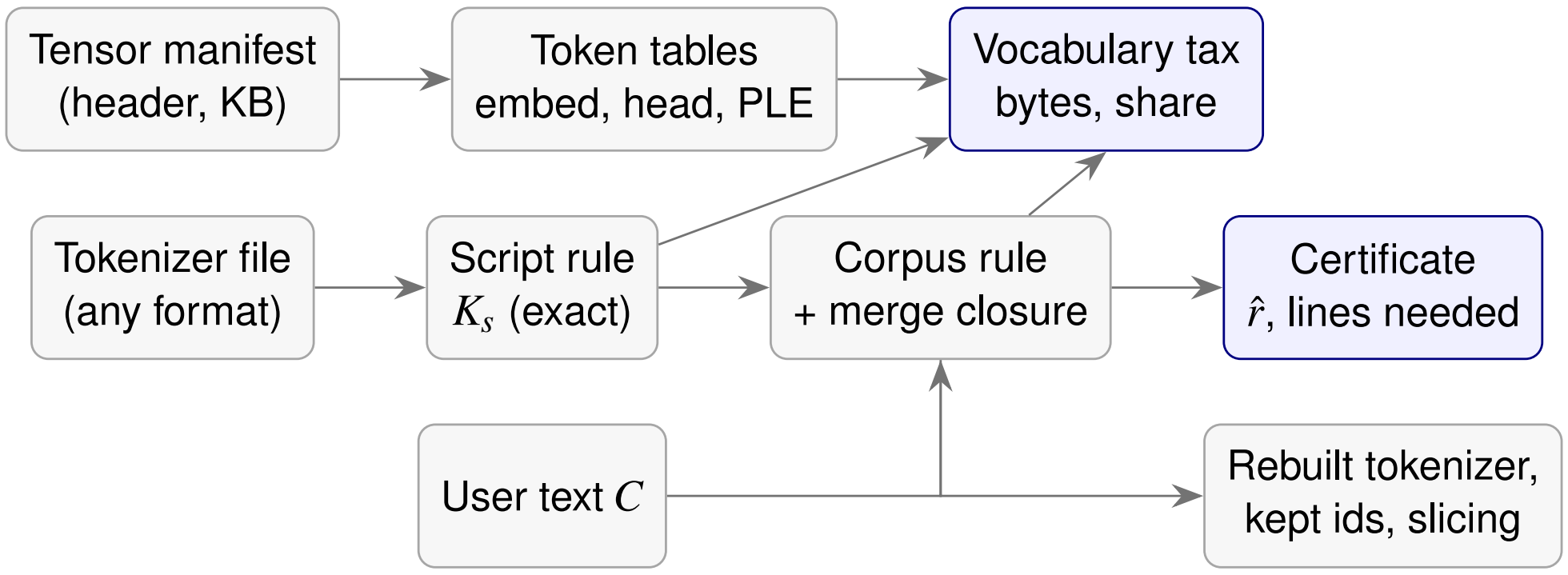

\section{Experimental setup}
\label{sec:setup}

\paragraph{Models.} We start from the 12,974 repositories of the Hub census of~\citet{singkul2026manifest}, whose token tables and stored sizes come from manifests. We assign each repository's tokenizer to one of 24 reference tokenizer families by model type and vocabulary size, using the vocabulary files distributed with llama.cpp~\citep{llamacpp}, the Whisper assets~\citep{radford2023whisper}, and the \tn{o200k\_base} and \tn{cl100k\_base} encodings used by GPT-OSS and OLMo. This maps \NMapped{} repositories. Of these, \NGen{} are generative or speech-recognition models (pipeline tags text generation, image-text-to-text, any-to-any and automatic speech recognition); we call them population P1. P1 is recent: 55\% of its models were created in 2025 or 2026, including Gemma~4, Qwen3, Ministral~3, GPT-OSS and OLMo~3. The newest families whose tokenizers are distributed only through the Hub (Qwen3.5 and later, Nemotron-H, GLM-5, DeepSeek-V3.2 and V4) are not yet mapped; they account for most of the downloads of models created in 2026, and the package computes their tax directly from their own tokenizer files. Population P2 is the \NCorpusPop{} models of P1 whose tokenizer belongs to one of the four BPE families we rebuild exactly from their merge lists (Gemma, Qwen, Llama~3 and GPT-2), 73\% of P1. Edge-sized models are those with fewer than 4B parameters.

\paragraph{Text.} For the script rule we use the 21 languages of the Parallel Universal Dependencies treebanks~\citep{demarneffe2021ud}, 1,000 parallel sentences each. For the corpus rule we take, for each of 22 languages, the largest Universal Dependencies treebank with sentence text (CC BY-SA or similar licenses), shuffle it, keep the first 1,000 sentences as held-out text and use the rest as trimming text; for Thai we add the Wisesight social-media corpus~\citep{wisesight}. We use up to 15,000 trimming lines per language. Held-out text comes from the same source as the trimming text, which is the situation a user who trims for their own domain is in.

\paragraph{Baselines.} We compare against the keep rules of prior work, each run on the same text and rebuilt the same way: the \emph{script rule} of \citet{bogoychev2024updowns} and SqueezeBits~\citep{squeezebits2025trim}; \emph{rank top-$K$}, which keeps the first $K$ pieces of the vocabulary ($K$ = 10k--80k); \emph{frequency top-$N$}, which keeps the $N$ most frequent pieces of the target text as in \citet{bogoychev2024updowns}, \citet{gurgurov2025compression} and the vocabulary part of COMPACT~\citep{kwek2025compact} ($N$ = 1k--16k); and \emph{seen in corpus}, the rule of \citet{ushio2023vt}, with minimum counts $m$ = 1--10. All rules keep special, byte and single-character pieces. Our method is the corpus rule with merge closure, with $m$ chosen by the certificate. For the certificate we compare against estimating the re-split rate on a 20\% hold-out of the trimming text and by 5-fold cross-validation, and for the extrapolation against fixed decay exponents.

\paragraph{Model quality.} We evaluate six edge-sized models with different tokenizers: Gemma~3 270M and 1B, Qwen3 0.6B and 1.7B, Llama~3.2 1B and SmolLM3 3B, the last two sharing one tokenizer. For each of 11 languages (Thai, Chinese, Japanese, Korean, Arabic, Hindi, Russian, German, French, Spanish, English) we compare the original model with the script rule and with the merge-closed corpus rule on 4,000 and 15,000 trimming lines (all available lines where a treebank has fewer: 3,997 for Chinese, 7,100 for Japanese, 6,526 for Arabic, 14,525 for English; Chinese has no 15,000-line condition). A trimmed model is evaluated exactly without editing weights: the trimmed tokenizer's ids are mapped to the original rows and every softmax runs over the kept rows only, which is the same function as a model whose embedding, head and per-layer tables are sliced. We measure bits per character on 300 held-out lines, which compares models with different tokenizations; the share of half-lines whose 24-token greedy continuation matches the original model's exactly; and zero-shot Belebele accuracy~\citep{bandarkar2024belebele} on 200 questions, choosing the answer with the lowest negative log-likelihood per character.

\paragraph{Tokenizers and the real re-split rate.} We rebuild the reference tokenizers with the Hugging Face \tn{tokenizers} and \tn{sentencepiece} libraries and check them against llama.cpp. For every trimmed keep set we rebuild the trimmed tokenizer and tokenize the held-out text with both; the \emph{real re-split rate} is the share of original tokens whose piece and character span do not appear in the trimmed tokenization, and the \emph{nominal} rate is the share of original tokens whose row was removed, the check used in prior work. Statistics are medians with 95\% bootstrap confidence intervals~\citep{efron1993} where intervals are shown.

\section{Results}
\label{sec:results}

\subsection{The tax is regressive and falls on small models}
\label{sec:rq1}

\begin{table}[t]
\centering\small
\caption{Reference tokenizer families, sorted by the share of vocabulary rows unused by Thai and English text under the script rule. Token ratio: Thai over English tokens for the same PUD sentence. $n$: generative and speech models of P1 using the family. Tax: their median share of model bytes for a Thai and English deployment.}
\label{tab:families}
\begin{tabular}{lrrrrr}
\toprule
Family & Vocabulary & Rows unused (\%) & Token ratio TH/EN & $n$ & Median tax (\%) \\
\midrule
Baichuan & 64{,}000 & 55.8 & 3.86 & 0 & -- \\
Aquila & 100{,}008 & 53.2 & 4.66 & 0 & -- \\
XLM-R & 250{,}048 & 51.9 & 1.06 & 0 & -- \\
Gemma 3/3n/4 & 262{,}144 & 35.0 & 1.53 & 170 & 5.3 \\
Qwen 2--3.5 & 151{,}936 & 28.8 & 2.36 & 496 & 4.1 \\
Command R & 256{,}000 & 28.1 & 4.31 & 1 & 1.7 \\
Mistral Tekken & 131{,}072 & 27.2 & 2.21 & 26 & 2.9 \\
GPT-OSS (o200k) & 199{,}998 & 27.1 & 1.84 & 17 & 4.5 \\
DeepSeek LLM & 102{,}400 & 24.8 & 7.51 & 8 & 0.6 \\
DeepSeek Coder & 32{,}256 & 20.5 & 7.97 & 0 & -- \\
Llama 3 & 128{,}256 & 15.7 & 2.04 & 205 & 2.1 \\
Whisper & 51{,}866 & 14.1 & 3.49 & 42 & 1.2 \\
Llama 2 & 32{,}000 & 13.2 & 4.01 & 169 & 0.5 \\
Phi-3 & 32{,}064 & 13.2 & 4.01 & 12 & 0.7 \\
Mistral v3 & 32{,}768 & 12.3 & 3.87 & 6 & 1.0 \\
StarCoder & 49{,}152 & 7.3 & 3.37 & 5 & 0.4 \\
BERT & 30{,}522 & 4.2 & 0.17 & 0 & -- \\
GPT-NeoX & 50{,}432 & 2.4 & 4.60 & 24 & 0.6 \\
Falcon & 65{,}024 & 2.4 & 7.54 & 7 & 0.2 \\
OLMo (cl100k) & 100{,}256 & 2.2 & 4.03 & 18 & 0.2 \\
GPT-2 & 50{,}257 & 0.5 & 8.29 & 38 & 0.1 \\

\bottomrule
\end{tabular}
\end{table}

\begin{figure}[t]
\centering
\includegraphics[width=0.8\linewidth]{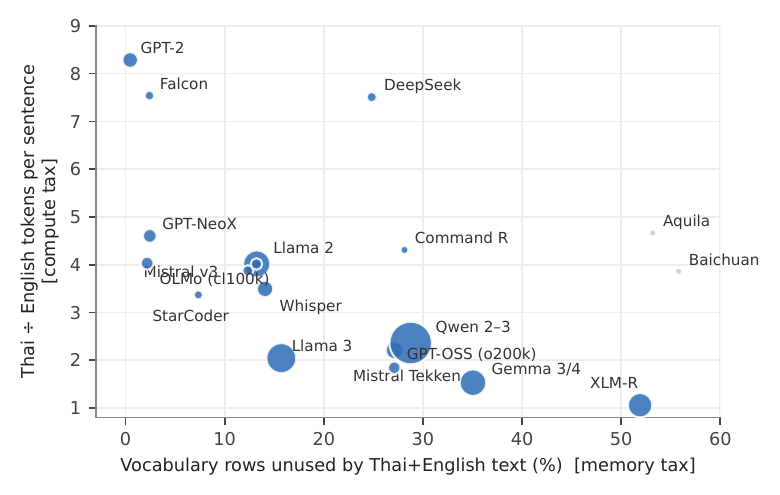}
\caption{The memory tax (rows a Thai and English deployment never uses) against the compute tax (Thai tokens per English token for the same sentence) for 19 tokenizer families. Dot area: models in P1 using the family; grey: none.}
\label{fig:twotax}
\end{figure}

Table~\ref{tab:families} lists the 24 tokenizer families. Between 0.5\% (GPT-2) and 56\% (Baichuan) of their rows are unused by Thai and English text. Families with large multilingual vocabularies carry the largest tax; families with small, English-centered vocabularies carry almost none, but split Thai into many more tokens. Figure~\ref{fig:twotax} plots the two costs against each other. GPT-2, Falcon and GPT-NeoX need 4.6--8.3 times as many tokens for Thai as for English and waste almost no memory; Gemma and XLM-R need 1.1--1.5 times as many tokens and leave 35--52\% of their rows unused, and the newest large vocabulary in the census, GPT-OSS's \tn{o200k}, sits between them (1.8 times, 27\%). No family in the census avoids both: a Thai user pays a compute tax or a memory tax, and the tokenizer decides which.

Across P1, the median tax for a Thai and English deployment is \TaxSA{} (95\% CI \TaxSciA) of model bytes for models below 0.5B parameters (n=\TaxSnA), \TaxSB{} for 0.5--2B, \TaxSC{} for 2--8B and \TaxSD{} above 8B (n=\TaxSnD). The 90th percentile below 0.5B is \TaxSpnA. Weighted by downloads, edge-sized models carry a tax of \TaxDLW{}. Across all of P1, which was downloaded \WasteDl{} million times in the month before the census, a median \WasteScript{} of the downloaded bytes over the 40 target languages (\WasteScriptLo{} for Chinese) are rows a single-language deployment never uses under the script rule, and \WasteCorpus{} in P2 under the Thai corpus rule; Qwen3 0.6B alone was downloaded 23 million times, each copy carrying 0.18\,GB of rows a Thai and English deployment never uses (0.54\,GB under the corpus rule). The decline is significant (Spearman $\rho=-0.34$ between parameters and tax over all \NGen{} models, $p<10^{-30}$; one-sided Mann--Whitney test of models under 0.5B against models over 8B, $p=4.6\times10^{-6}$). The token tables themselves hold a median 13.1\% of model bytes in P1 and 41.2\% below 0.5B. The tax is regressive for a structural reason: the token tables grow with the vocabulary and the width, while the transformer layers grow with the width squared times the depth, so the tables are a larger share of a small model. Per-layer embeddings make this extreme. In Gemma~4 E2B the token tables hold 2.75B of 5.12B parameters, 4.70\,GB of them in per-layer tables, and \ExGemmaFourSGB{}\,GB of its \ExGemmaFourGB{}\,GB (\ExGemmaFourS) are rows a Thai and English user never uses (Table~\ref{tab:models}); the 41 models of P1 with per-layer embeddings have a median tax of 15.3\%.

The tax barely depends on which language is kept. Repeating the census for 40 languages (Appendix Table~\ref{tab:app-script}), the median below 0.5B is 4.4--4.5\% for 35 of them; it is lower only for languages whose script occupies a large part of common vocabularies (Korean 4.0\%, Chinese 3.6\%, Japanese 3.3\%, Russian and Ukrainian 2.4\%). For Lao and Punjabi the script rule keeps no more rows than for English alone, and for Amharic, Burmese and Khmer hardly more: the reference vocabularies contain hardly any pieces in these scripts, so their speakers pay the compute tax in full instead.

\subsection{The script rule is exact}
\label{sec:rq2}

Over 189 language--tokenizer pairs (21 PUD languages, 9 families), at most \MissMax{} of held-out tokens fall in rows the script rule removes, and the misses are rare symbols outside any declared script (Appendix Table~\ref{tab:app-valid}). Table~\ref{tab:surgery} reports the full round trip for five families: we rebuild each tokenizer, trim it to Thai and English, and tokenize every PUD sentence and 4,000 Thai social-media posts with both. Thai and the 14 Latin-script languages, English among them, tokenize identically in 100\% of sentences, every language round-trips to the same text, and the removed scripts still encode, at 1.0--7.2 times as many tokens. The rebuilt untrimmed tokenizers agree with llama.cpp on all sentences for Qwen, GPT-2 and Gemma and on 97.9\% for Llama~3, where they differ in pre-tokenization of a few character classes; the identity test compares the trimmed and untrimmed tokenizers built the same way.

\begin{table}[t]
\centering\small
\caption{Tokenizer surgery to Thai and English; all columns but the last in \%. Rebuild: share of sentences on which the rebuilt untrimmed tokenizer matches llama.cpp. Same: share of sentences tokenized identically by the trimmed and untrimmed tokenizer (Thai: PUD and 4,000 social-media posts; Latin: 14 Latin-script PUD languages). Round trip: share of all sentences decoded back to the input. Other scripts: token count after trimming over before, for Arabic, Hindi, Japanese, Korean, Russian and Chinese. Mistral v3 is a SentencePiece model, trimmed in its protocol buffer and compared with the \tn{sentencepiece} library only.}
\label{tab:surgery}
\begin{tabular}{lrrrrrr}
\toprule
Family & Kept / vocabulary & Rebuild & Same TH & Same Latin & Round trip & Other scripts \\
\midrule
Qwen 2--3.5 & 108{,}232\,/\,151{,}936 & 100.0 & 100 & 100 & 100 & 1.30--4.97 \\
Llama 3 & 108{,}101\,/\,128{,}256 & 97.9 & 100 & 100 & 100 & 2.04--5.43 \\
GPT-2 & 50{,}021\,/\,50{,}257 & 100.0 & 100 & 100 & 100 & 1.00--1.63 \\
Gemma 3/3n/4 & 170{,}274\,/\,262{,}144 & 100.0 & 100 & 100 & 100 & 3.62--7.23 \\
Mistral v3 & 28{,}726\,/\,32{,}768 & -- & 100 & 100 & 100 & 1.91--4.65 \\

\bottomrule
\end{tabular}
\end{table}

Keeping Latin matters. Trimming to the Thai script alone removes 91--95\% of the rows instead of 12--35\% (0.5\% for GPT-2), but 3.5--6.6\% of the tokens in Thai text are then lost (1.8\% for GPT-2), because Thai text contains Latin names, numbers written with Latin pieces, URLs and English words. With Latin kept, no Thai token is lost.

\subsection{Frequency-based trimming breaks BPE tokenizers}
\label{sec:closure}

\begin{figure}[t]
\centering
\includegraphics[width=0.95\linewidth]{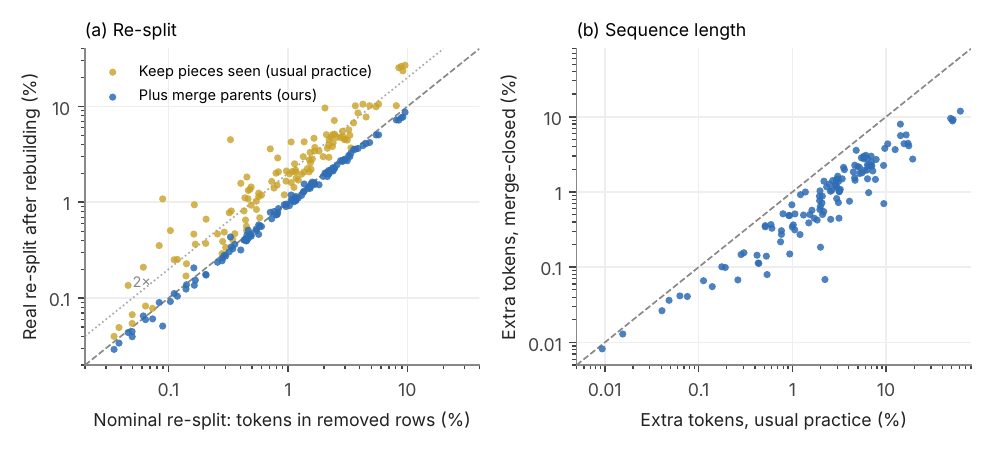}
\caption{Corpus trimming with and without merge closure, 22 languages $\times$ 4 BPE families, 4,000 trimming lines. (a) Real re-split rate of 1,000 held-out lines after rebuilding the tokenizer, against the nominal rate (tokens whose row was removed). Keeping the pieces seen as tokens (gold) gives a real rate a median \NaiveRatio{} times the nominal rate; adding merge parents (blue) brings it to or below the nominal rate. (b) Extra tokens per held-out text, without and with closure.}
\label{fig:closure}
\end{figure}

\begin{table}[t]
\centering\small
\caption{Merge closure, medians over 22 languages per family (4,000 trimming lines, 2,000 for Chinese and Urdu). Orphans: kept rows the rebuilt tokenizer can no longer produce without closure. Closure cost: extra kept rows. Real / nominal: real re-split rate over nominal rate, without and with closure (settings with a nominal rate above 0.2\%). Extra tokens: median / maximum growth of held-out sequences (\%).}
\label{tab:closure}
\begin{tabular}{lrrrrrr}
\toprule
 & & Closure & \multicolumn{2}{c}{Real / nominal} & \multicolumn{2}{c}{Extra tokens (\%)} \\
\cmidrule(lr){4-5}\cmidrule(lr){6-7}
Family & Orphans & cost (\%) & usual & closed & usual & closed \\
\midrule
Gemma 3/3n/4 & 1{,}146 & 8.7 & 2.05 & 0.95 & 6.31 / 62 & 1.80 / 12 \\
Qwen 2--3.5 & 989 & 4.7 & 2.10 & 0.95 & 2.86 / 52 & 0.72 / 9 \\
Llama 3 & 638 & 11.6 & 1.81 & 0.91 & 2.07 / 49 & 0.88 / 10 \\
GPT-2 & 567 & 8.1 & 1.67 & 0.94 & 0.76 / 51 & 0.36 / 9 \\

\bottomrule
\end{tabular}
\end{table}

Figure~\ref{fig:closure} and Table~\ref{tab:closure} compare the usual corpus rule, which keeps the pieces seen as tokens, with its merge-closed version, measuring both on rebuilt tokenizers. Without closure, between \OrphMin{} and \OrphMax{} kept rows per setting are orphans (median \OrphMed, \OrphShare{} of the kept rows): they cost memory and can never be selected. The real re-split rate exceeds the nominal rate in almost every setting, by a median factor of \NaiveRatio{} and by more than 1.2 in \NaiveWorse{} of the \NaiveMeas{} settings with a measurable rate. The effect is largest for Latin-script text, where long words are built by chains of merges: for German on Qwen the nominal rate is 2.4\% and the real rate 7.1\%; for English web text the real rate is 23--27\% and held-out sequences grow by 49--62\%, against a nominal rate of about 9\%. With closure the real rate is a median \ClosedRatio{} of the nominal rate (below 1 because the added parents also catch some words that the removed pieces would have covered), and sequences grow by a median \ClosedTok{} instead of \NaiveTok. Closure costs a median \ExtraRows{} more kept rows (up to \ExtraRowsMax{} for English), a small price next to the 80--90\% of rows the corpus rule removes. The script rule needs almost no closure: on Qwen with Thai and English it adds 92 parents to 108,232 kept pieces: single Cyrillic, Arabic, Hangul and CJK characters that are parents of byte-fragment pieces, which the rule always keeps.

\subsection{Merge-closed trimming beats prior keep rules}
\label{sec:baselines}

\begin{figure}[t]
\centering
\includegraphics[width=\linewidth]{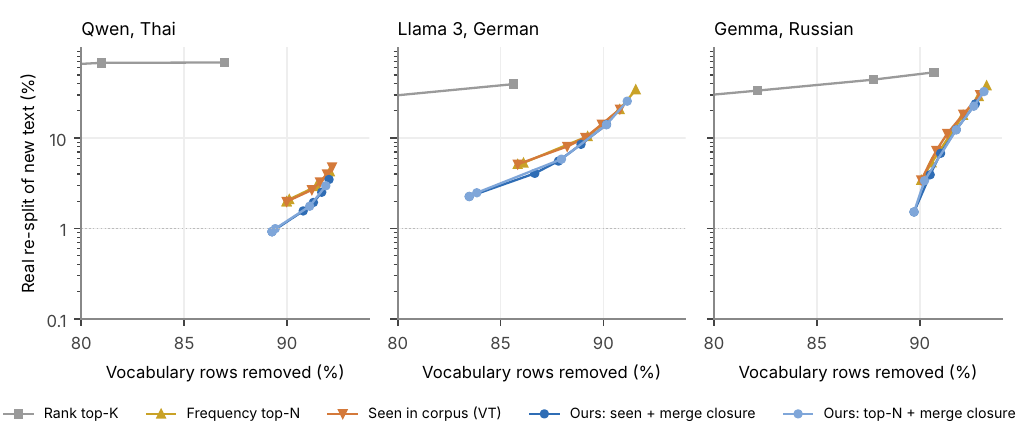}
\caption{Rows removed against the real re-split rate of new text for each keep rule, swept over its parameter (4,000 trimming lines, rebuilt tokenizers). Lower and to the right is better. Merge closure moves both corpus-based rules down and to the left of their usual versions; rank top-$K$ keeps the wrong pieces altogether.}
\label{fig:pareto}
\end{figure}

\begin{table}[t]
\centering\small\setlength{\tabcolsep}{4pt}
\caption{Keep rules compared on 44 settings (4 families $\times$ 11 languages: Thai, Chinese, Japanese, Korean, Arabic, Hindi, Russian, German, French, Spanish, English; 4,000 trimming lines). Matched: real re-split rate of the rule at the rows our rule removes, interpolated along its curve, as a multiple of ours (median). Orphans: unreachable kept rows (median). Oracle: mean rows removed (\%) at a real re-split rate $\le$1\%, choosing each rule's best parameter with knowledge of the real rate, script rule as fallback. Deployed: each rule chooses its parameter as a user would, prior rules by the nominal rate on 1,000 held-out lines (test data), ours by the certificate on the trimming text alone (no test data); mean rows removed (\%) and number of settings whose real re-split rate exceeds the target.}
\label{tab:baselines}
\begin{tabular}{lrrrrrrr}
\toprule
 & & & Oracle & \multicolumn{2}{c}{Deployed, $\tau$=1\%} & \multicolumn{2}{c}{Deployed, $\tau$=2\%} \\
\cmidrule(lr){5-6}\cmidrule(lr){7-8}
Keep rule & Matched & Orphans & $\tau$=1\% & Removed & Over $\tau$ & Removed & Over $\tau$ \\
\midrule
Script rule~\citep{bogoychev2024updowns,squeezebits2025trim} & -- & 87 & 17.3 & 17.3 & 0 & 17.3 & 0 \\
Rank top-$K$ (first $K$ pieces) & 8.4$\times$ & 87 & 22.4 & 22.4 & 0 & 24.0 & 0 \\
Frequency top-$N$~\citep{bogoychev2024updowns,gurgurov2025compression,kwek2025compact} & 1.32$\times$ & 424 & 46.5 & 51.1 & 3 & 68.1 & 9 \\
Seen in corpus~\citep{ushio2023vt} & 1.30$\times$ & 720 & 46.6 & 51.2 & 6 & 68.5 & 13 \\
\textbf{Ours} (merge-closed, certificate) & 1$\times$ & 0 & \textbf{52.6} & \textbf{52.6} & 3 & 68.2 & \textbf{0} \\

\bottomrule
\end{tabular}
\end{table}

Figure~\ref{fig:pareto} and Table~\ref{tab:baselines} compare our rule with the keep rules of prior work. At the same number of rows removed, the seen-in-corpus rule re-splits a median \MatchVT{} times as much new text as ours and frequency top-$N$ \MatchFreq{} times as much, worse in \MatchShareVT{} and \MatchShareFreq{} of the comparisons; rank top-$K$, which ignores the target language, re-splits \MatchRank{} times as much. Choosing each rule's best parameter with knowledge of the real rate, ours removes \OrcOurs{} of the rows on average at a real rate of at most 1\%, against \OrcVT{} for seen-in-corpus, \OrcFreq{} for frequency top-$N$, \OrcRank{} for rank top-$K$ and \OrcScript{} for the script rule alone.

The deployed comparison is the one a user faces. A user of a prior rule checks the nominal rate on held-out text and keeps the most aggressive setting that passes; because the nominal rate understates the real one, this setting overshoots the target in \ViolVTtwo{} of 44 settings for seen-in-corpus and \ViolFreqtwo{} for frequency top-$N$ at a 2\% target, although these users had test data. Our user, without test data, chooses by the certificate and overshoots in \ViolOurstwo{} settings while removing about as many rows (68.2\% against 68.1--68.5\%). At a 1\% target the certificate overshoots in \ViolOursone{} settings, as many as frequency top-$N$ (\ViolFreqone) and half as many as seen-in-corpus (\ViolVTone); our overshoots reach at most 1.18\%, those of the prior rules up to 1.58\% and 2.53\%. Every prior rule also keeps unreachable rows, the script rule included (a median of 87 byte-fragment pieces whose parents it removes); ours keeps none.

\subsection{Every language can shed the same share, at very different cost}
\label{sec:rq3}

\begin{table}[t]
\centering\small
\caption{Merge-closed corpus rule for Thai: share of vocabulary rows removed / real re-split rate of 1,000 held-out lines (\%), by number of trimming lines. Script: rows removed by the script rule alone.}
\label{tab:thai}
\begin{tabular}{lrrrrr}
\toprule
Family & Script & 1,000 & 2,000 & 4,000 & 15,000 \\
\midrule
Gemma 3/3n/4 & 35.0 & 92.1 / 3.33 & 91.4 / 2.35 & 90.7 / 1.59 & 88.5 / 0.67 \\
Qwen 2--3.5 & 28.8 & 90.9 / 2.00 & 90.0 / 1.46 & 89.3 / 0.92 & 86.5 / 0.40 \\
Llama 3 & 15.7 & 89.5 / 2.28 & 88.3 / 1.65 & 87.3 / 1.05 & 83.5 / 0.44 \\
GPT-2 & 0.5 & 90.5 / 0.58 & 88.4 / 0.41 & 86.3 / 0.28 & 79.1 / 0.10 \\

\bottomrule
\end{tabular}
\end{table}

\begin{figure}[t]
\centering
\includegraphics[width=0.7\linewidth]{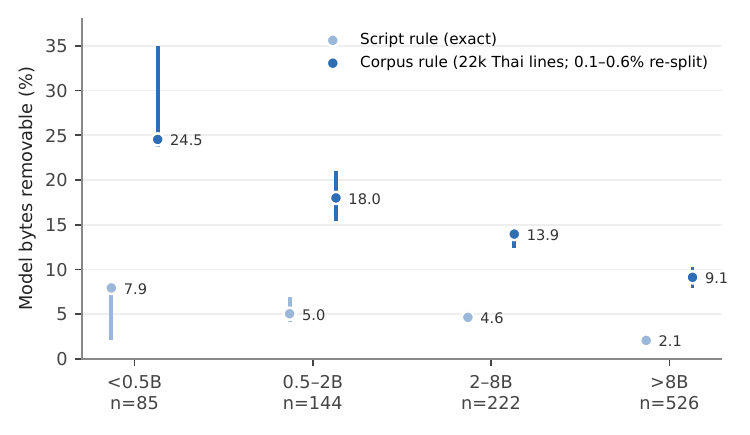}
\caption{The tax is regressive. Median share of model bytes removable for a Thai and English deployment by model size in P2, under the script rule and the merge-closed corpus rule with 15,000 Thai lines. Bars: 95\% bootstrap confidence intervals.}
\label{fig:regressive}
\end{figure}

With 15,000 lines of Thai text, the merge-closed corpus rule removes \ThRemLo--\ThRemHi{} of the rows of the four families at a real re-split rate of \ThRsLo--\ThRsHi{} (Table~\ref{tab:thai}); each doubling of the text lowers the rate by about a third. In P2 the median tax rises from \TaxCsA{} under the script rule to \TaxCA{} below 0.5B parameters, and from \TaxCsD{} to \TaxCD{} above 8B (Figure~\ref{fig:regressive}). For Gemma~4 E2B the corpus rule removes \ExGemmaFourCGB{}\,GB of \ExGemmaFourGB{}\,GB (\ExGemmaFourC), and for Qwen3 0.6B \ExQwenSmallC{} (Table~\ref{tab:models}).

\begin{figure}[t]
\centering
\includegraphics[width=0.9\linewidth]{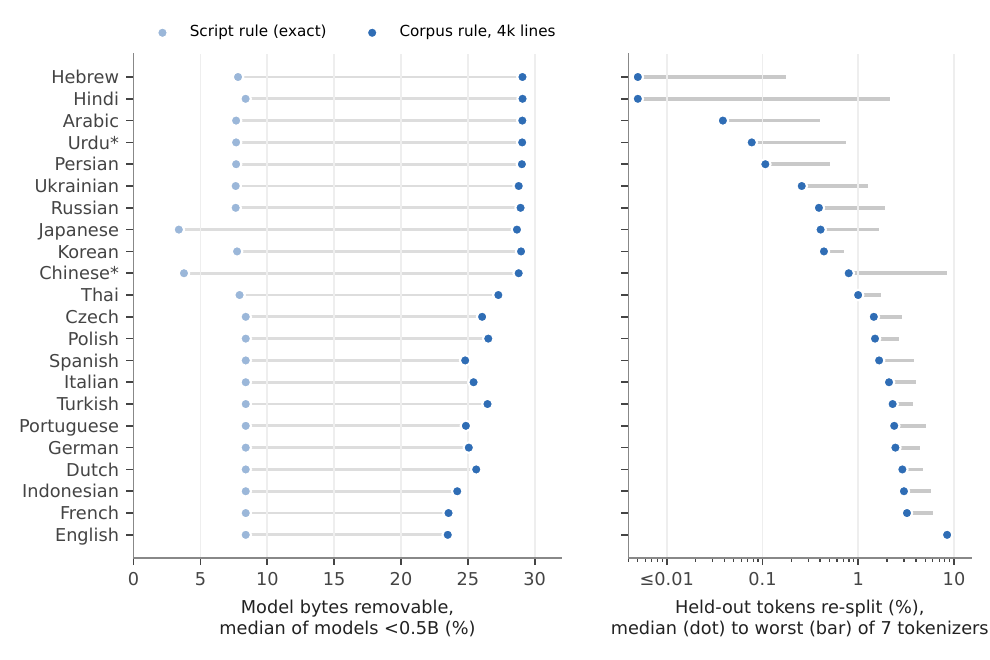}
\caption{22 languages, P2 (\NCorpusPop{} models), merge-closed corpus rule with 4,000 trimming lines (Urdu and Chinese: 2,000). Left: median share of bytes removable for models under 0.5B. Right: real re-split rate of new text, median and range of the four families, log scale. Languages are sorted by the median.}
\label{fig:multilingual}
\end{figure}

Figure~\ref{fig:multilingual} repeats this for 22 languages. The share of bytes removable below 0.5B is \MLCorpLo--\MLCorpHi{} in every language: each language uses a small fraction of a large vocabulary. The cost of removing it is not equal. With 4,000 lines the median real re-split rate is at most 0.01\% for Hebrew and Hindi, \RSAr{} for Arabic, \RSRu{} for Russian and \RSTh{} for Thai, but \RSDe{} for German, \RSFr{} for French and \RSEn{} for English web text. The order follows how much of a language's vocabulary lies in pieces shared with English. Pieces in a script used by few languages are few and frequent, so a few thousand lines see nearly all of those a language needs. Latin pieces are numerous and spread over every Latin-script language, so a sample of German sees a smaller share of the German pieces it will need; English, whose pieces are the bulk of most vocabularies, is the extreme. Appendix Table~\ref{tab:app-corpus} gives every language and family, with and without closure.

\subsection{Mixed-language deployments}
\label{sec:mix}

\begin{table}[t]
\centering\small\setlength{\tabcolsep}{4pt}
\caption{Deployments serving several languages (ISO codes; English always included). Script rule: median tax (\%) of models under 0.5B in P1, and GB removable from Gemma~4 E2B. Corpus rule: merge-closed, 2,000 trimming lines per language, medians over the four families of rows removed (\%) and pooled real re-split rate (\%), the worst real rate of any single language and family (\%), and the median ratio of the per-language certificate to the per-language real rate.}
\label{tab:mix}
\begin{tabular}{lrrrrrr}
\toprule
 & \multicolumn{2}{c}{Script rule} & \multicolumn{4}{c}{Merge-closed corpus rule} \\
\cmidrule(lr){2-3}\cmidrule(lr){4-7}
Languages & Tax $<$0.5B & Gemma 4 E2B & Removed & Real (pooled) & Worst language & Cert./real \\
\midrule
th, zh, en & 3.3 & 1.52 & 85.7 & 1.93 & 7.08 & 0.98 \\
ar, fr, en & 4.4 & 1.80 & 83.2 & 1.86 & 8.08 & 1.04 \\
ja, ko, zh, en & 2.7 & 1.37 & 85.3 & 1.37 & 6.66 & 0.96 \\
es, pt, fr, en & 4.5 & 1.97 & 79.9 & 1.99 & 6.26 & 1.09 \\
ru, uk, en & 2.4 & 1.69 & 89.7 & 0.47 & 2.91 & 1.11 \\

\bottomrule
\end{tabular}
\end{table}

Most deployments serve more than one language. Table~\ref{tab:mix} repeats the analysis for five mixtures, each with English. Under the script rule the tax shrinks when the mixture includes a script that occupies much of the vocabulary: Thai with Chinese leaves a median 3.3\% below 0.5B and 1.52\,GB in Gemma~4 E2B, against 4.4\% and 1.93\,GB for Thai alone. The merge-closed corpus rule still removes 80--90\% of the rows with 2,000 lines per language. The pooled re-split rate hides large differences between the languages of a mixture (French reaches 8.1\% next to Arabic's 0.5\% on Gemma), so the certificate should be read per language. We compute it per language by counting, in each language's text, the singletons that no other language of the mixture uses; it predicts the per-language real rate with a median ratio of \MixPer{} (interquartile range \MixPerIQR, \MixPerN{} language--family pairs), and the pooled estimate predicts the pooled rate with a median ratio of \MixPool. Without merge closure the pooled real rate is \MixNaive{} times higher.

\subsection{The certificate predicts the real re-split rate without test data}
\label{sec:cert}

\begin{figure}[t]
\centering
\includegraphics[width=0.95\linewidth]{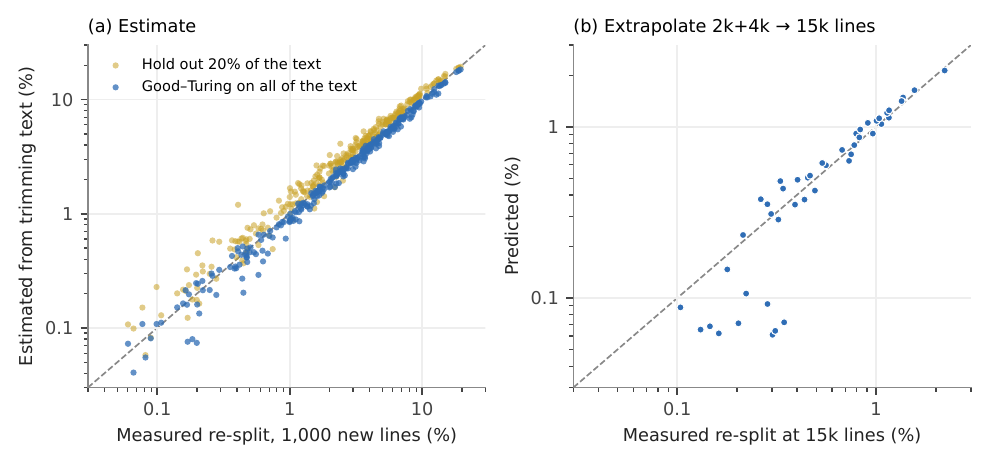}
\caption{Trim certificate, against the real re-split rate of merge-closed trimming. (a) Rate estimated from the trimming text (Eq.~\ref{eq:gt}, blue) or measured on a 20\% hold-out of it (gold), against the real rate on 1,000 new lines; 22 languages, 4 families, 1k, 2k and 4k lines. (b) Rate at 15,000 lines predicted from 2,000 and 4,000 lines (Eq.~\ref{eq:need}) against the real rate. Dashed: perfect agreement.}
\label{fig:cert}
\end{figure}

We evaluate the certificate on 22 languages, the four families and 1k, 2k, 4k and, where the treebank allows, 15k trimming lines: \CertN{} settings, each against the real re-split rate of the merge-closed trimmed tokenizer on the language's 1,000 held-out lines. The estimate of Eq.~\ref{eq:gt} agrees with the real rate with correlation \CertR{} on a log scale and a median ratio of \CertRatio{} (interquartile range \CertIQR), and it falls on the right side of a 1\% threshold in \CertAgree{} of \CertN{} settings (Figure~\ref{fig:cert}a). Holding out 20\% of the trimming text, the alternative that needs no theory, has a median relative error of \CertSpA, \CertSpB{} and \CertSpC{} at 1k, 2k and 4k lines, against \CertGTA, \CertGTB{} and \CertGTC{} for the estimate. Five-fold cross-validation, which uses all of the text for estimation, does no better (median error \CVerr; Table~\ref{tab:certbase}): every fold builds its keep set from 80\% of the text, while the deployed set uses all of it, so both resampling estimates are biased upward. The Good--Turing estimate has no such bias because it describes the keep set actually deployed.

\begin{table}[t]
\centering\small
\begin{minipage}[t]{0.52\linewidth}\centering
\captionof{table}{Median relative error (\%) of re-split estimates against the real rate, by trimming lines. $n$: settings with a real rate above 0.2\%.}
\label{tab:certbase}
\begin{tabular}{rrrrr}
\toprule
Lines & Hold-out & 5-fold CV & \textbf{Ours} & $n$ \\
\midrule
1{,}000 & 22.6 & 24.9 & \textbf{6.6} & 76 \\
2{,}000 & 24.4 & 25.5 & \textbf{5.9} & 72 \\
4{,}000 & 25.9 & 26.7 & \textbf{4.5} & 62 \\
15{,}000 & -- & 25.8 & \textbf{6.5} & 32 \\

\bottomrule
\end{tabular}
\end{minipage}\hfill
\begin{minipage}[t]{0.45\linewidth}\centering
\captionof{table}{Predicting the real rate at 15k lines from 2k and 4k lines (35 pairs): median ratio to the real rate and median $|\log|$ error.}
\label{tab:extbase}
\begin{tabular}{lrr}
\toprule
Predictor & Ratio & $|\log|$ err. \\
\midrule
No change (rate at 4k lines) & 2.75 & 1.01 \\
Fixed exponent $\beta=0.5$ & 1.42 & 0.39 \\
Fixed exponent $\beta=1$ & 0.73 & 0.32 \\
\textbf{Fitted $\beta$ (Eq.~\ref{eq:need})} & 1.03 & 0.10 \\

\bottomrule
\end{tabular}
\end{minipage}
\end{table}

To test the extrapolation of Eq.~\ref{eq:need}, we predict the real rate at 15,000 lines from the estimates at 2,000 and 4,000 lines for the 11 languages whose treebanks are large enough (\ExtN{} language--family pairs with a rate above 0.1\%). The median ratio of predicted to real rate is \ExtRatio{} (interquartile range \ExtIQR; \ExtRatioHi{} for the \ExtNHi{} pairs with a rate of at least 0.3\%). A fixed exponent does markedly worse ($|\log|$ error \ExtBoneErr{} for $\beta=1$ against \ExtFitErr{} fitted; Table~\ref{tab:extbase}), so the decay must be fitted per language and tokenizer. Below 0.3\% the prediction can be low by up to a factor of four (Figure~\ref{fig:cert}b), because the estimate becomes noisy when few singletons remain; at a 1\% target this region is already certified. Table~\ref{tab:lines} turns the extrapolation into a guide: the median number of lines needed for a 1\% rate is about 4,000 or fewer for the languages written in Arabic, Hebrew, Devanagari, Korean, Cyrillic or Thai script, 5,000 for Japanese, 7,000--20,000 for the Latin-script languages, and about 190,000 for English web text.

\begin{table}[t]
\centering\small
\caption{Lines of in-domain text needed for a real re-split rate of at most 1\% after merge-closed trimming, extrapolated from 2,000 and 4,000 lines (Eq.~\ref{eq:need}); median and worst of the four families. Measured: median real rate at 4,000 lines (\%).}
\label{tab:lines}
\begin{tabular}{lrrr}
\toprule
Language & Measured at 4k (\%) & Lines needed (median) & Lines needed (worst) \\
\midrule
Arabic & 0.04 & $\le$4k & $\le$4k \\
Persian & 0.14 & $\le$4k & $\le$4k \\
Hebrew & 0.00 & $\le$4k & $\le$4k \\
Hindi & 0.00 & $\le$4k & 8k \\
Korean & 0.41 & $\le$4k & $\le$4k \\
Russian & 0.39 & $\le$4k & 6k \\
Ukrainian & 0.40 & $\le$4k & 4k \\
Thai & 0.99 & 4k & 7k \\
Japanese & 0.79 & 5k & 6k \\
Spanish & 1.56 & 7k & 16k \\
Polish & 1.54 & 7k & 13k \\
Czech & 1.76 & 9k & 18k \\
Italian & 1.95 & 10k & 20k \\
Portuguese & 2.21 & 12k & 27k \\
German & 2.26 & 14k & 25k \\
Turkish & 2.42 & 15k & 25k \\
Indonesian & 2.74 & 18k & 55k \\
Dutch & 2.78 & 19k & 35k \\
French & 3.04 & 20k & 43k \\
English (web) & 7.69 & 191k & 275k \\

\bottomrule
\end{tabular}
\end{table}

\paragraph{Shifting domain.} The certificate describes new text from the source of the trimming text. To measure what happens when the deployment text differs, we trim on each language's main treebank (for Thai, on social-media posts only) and test on the PUD sentences of the same language, translated news and Wikipedia text (Table~\ref{tab:domain}). Across the 16 languages the real rate on the shifted domain is a median \DomRatio{} times the in-domain rate and up to \DomRatioMax{} times (Korean); it is roughly unchanged for French, Indonesian and Japanese. Adding 200 lines of the new domain to the trimming text lowers it by only \DomAdd. Thai moves the other way, from 1.47\% to 0.12\%, because informal posts mix in English, slang and spelling variants that make formal news easy to cover. The practical remedy is cheap because of merge closure: with a closed keep set, the nominal rate measured on just 200 lines of deployment text is a median \DomNom{} of the real rate on 1,000 lines, so a user can check a new domain directly; without closure the same check would understate the damage about twofold (Section~\ref{sec:closure}).

\begin{table}[t]
\centering\small
\caption{Domain shift, 4,000 trimming lines from the main treebank (Thai: social-media posts), medians over four families. Certificate and real re-split rate on 1,000 in-domain lines and on the 1,000 PUD sentences (\%), their ratio, and the nominal rate on 200 PUD lines over the real rate on all 1,000. Chinese is omitted because its PUD is in Traditional characters.}
\label{tab:domain}
\begin{tabular}{lrrrrr}
\toprule
Language & Certificate & Real, in-domain & Real, PUD & PUD / in-domain & Nominal 200 / real \\
\midrule
Arabic & 0.06 & 0.04 & 0.15 & -- & 1.22 \\
Czech & 1.87 & 1.76 & 3.14 & 1.75 & 0.77 \\
English & 8.95 & 7.69 & 13.65 & 1.78 & 0.82 \\
French & 3.10 & 3.04 & 2.96 & 0.98 & 0.96 \\
German & 2.35 & 2.26 & 4.74 & 2.10 & 0.78 \\
Hindi & 0.00 & 0.00 & 0.10 & 2.13 & 0.85 \\
Indonesian & 2.52 & 2.74 & 2.77 & 1.00 & 0.84 \\
Italian & 2.08 & 1.95 & 3.69 & 1.89 & 0.89 \\
Japanese & 0.88 & 0.79 & 0.88 & 1.12 & 1.05 \\
Korean & 0.24 & 0.41 & 1.26 & 2.79 & 1.26 \\
Polish & 1.49 & 1.54 & 3.69 & 2.34 & 0.82 \\
Portuguese & 2.33 & 2.21 & 3.51 & 1.59 & 0.95 \\
Russian & 0.40 & 0.39 & 0.86 & 2.10 & 1.12 \\
Spanish & 1.61 & 1.56 & 3.09 & 1.99 & 0.83 \\
Thai & 1.01 & 1.47 & 0.12 & 0.08 & 0.50 \\
Turkish & 2.54 & 2.42 & 4.27 & 1.75 & 0.72 \\

\bottomrule
\end{tabular}
\end{table}

\paragraph{What did not help.} A natural way to need less text is to keep, in addition to the pieces seen, the first $K$ pieces of the vocabulary, since BPE merge order follows pretraining frequency. With $K=16{,}000$ this lowers the nominal re-split rate of Qwen on 1,000 French lines from \RPa{} to \RPb{} but keeps \RPcost{} percentage points more rows, a worse trade than adding text (Appendix~\ref{app:ablations}). We keep the plain corpus rule and let the certificate say how much text is needed.

\subsection{Trimmed models keep their predictions}
\label{sec:quality}

\begin{table}[t]
\centering\small
\caption{Quality of trimmed models, over the 11 languages of each condition (n). Rows removed and measured re-split rate on the evaluation lines: medians. Change in bits per character (BPC): median and range. Greedy: share of greedy continuations identical to the original model's, mean. Belebele: mean change in accuracy, percentage points (original accuracy 0.20--0.39, chance 0.25).}
\label{tab:quality}
\setlength{\tabcolsep}{4.5pt}
\begin{tabular}{llrrrrrrr}
\toprule
Model & Rule & n & Removed & Re-split & $\Delta$BPC & Range & Greedy & $\Delta$Belebele \\
 & & & (\%) & (\%) & (\%) & (\%) & (\%) & (pp) \\
\midrule
Gemma 3 270M & Script & 11 & 34.1 & 0.00 & $-$0.01 & $-$0.1, $+$0.4 & 100 & 0.0 \\
 & Corpus 4k & 11 & 88.5 & 1.58 & $+$2.93 & $-$0.2, $+$23.3 & 63 & $+$0.2 \\
 & Corpus 15k & 10 & 87.9 & 0.80 & $+$1.13 & $-$0.3, $+$8.1 & 80 & $+$0.2 \\
\addlinespace[2pt]
Qwen3 0.6B & Script & 11 & 28.9 & 0.00 & $-$0.20 & $-$0.4, $+$0.1 & 100 & 0.0 \\
 & Corpus 4k & 10 & 89.1 & 0.95 & $+$0.54 & $-$1.9, $+$14.0 & 76 & $+$0.5 \\
 & Corpus 15k & 7 & 86.7 & 0.42 & $-$1.04 & $-$1.8, $+$0.4 & 85 & $+$0.3 \\
\addlinespace[2pt]
Gemma 3 1B & Script & 11 & 34.1 & 0.00 & $-$0.07 & $-$0.6, $+$0.6 & 99 & 0.0 \\
 & Corpus 4k & 11 & 87.6 & 1.74 & $+$2.96 & $-$0.2, $+$21.3 & 53 & $-$0.5 \\
 & Corpus 15k & 10 & 87.6 & 0.81 & $+$1.15 & $-$0.6, $+$9.2 & 75 & $-$0.2 \\
\addlinespace[2pt]
Llama 3.2 1B & Script & 11 & 15.6 & 0.00 & $-$0.03 & $-$0.3, $+$0.3 & 100 & 0.0 \\
 & Corpus 4k & 11 & 88.0 & 0.60 & $+$0.69 & $-$2.2, $+$16.7 & 72 & $-$1.4 \\
 & Corpus 15k & 10 & 85.4 & 0.35 & $-$0.49 & $-$2.2, $+$4.5 & 85 & $-$0.8 \\
\addlinespace[2pt]
Qwen3 1.7B & Script & 11 & 28.9 & 0.00 & $-$0.39 & $-$0.7, 0.0 & 99 & 0.0 \\
 & Corpus 4k & 11 & 88.8 & 0.98 & $+$2.27 & $-$0.9, $+$16.3 & 72 & $-$0.2 \\
 & Corpus 15k & 10 & 87.5 & 0.47 & $-$0.13 & $-$0.9, $+$4.5 & 86 & $+$0.3 \\
\addlinespace[2pt]
SmolLM3 3B & Script & 11 & 15.6 & 0.00 & $-$0.02 & $-$0.1, $+$0.3 & 100 & 0.0 \\
 & Corpus 4k & 11 & 88.0 & 0.60 & $-$8.75 & $-$13.8, $+$14.2 & 72 & $-$2.0 \\
 & Corpus 15k & 10 & 85.4 & 0.35 & $-$8.37 & $-$13.8, $+$3.8 & 83 & $-$0.9 \\

\bottomrule
\end{tabular}
\end{table}

\begin{figure}[t]
\centering
\includegraphics[width=0.95\linewidth]{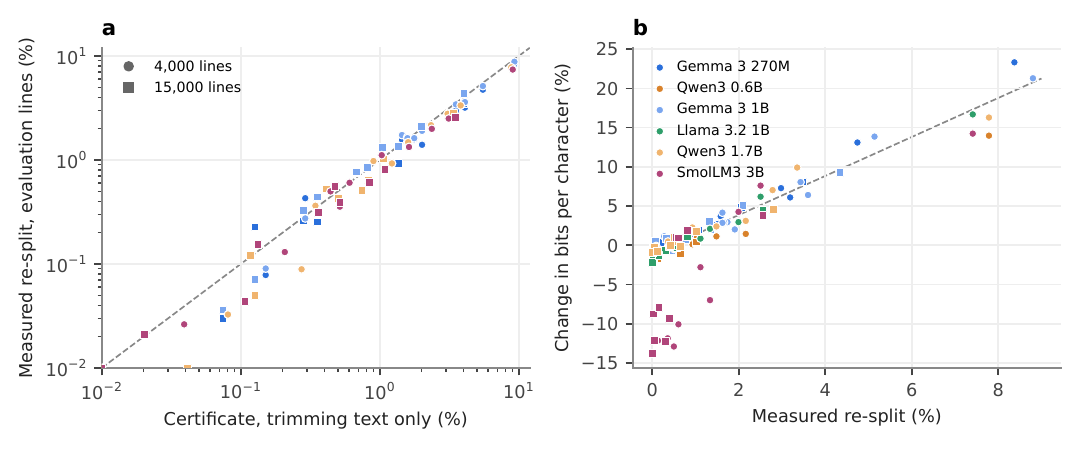}
\caption{Corpus rule on six models and 11 languages (\QCorpN{} settings). (a) The certificate, computed from the trimming text, against the re-split rate measured on the evaluation lines (correlation \QCertR). Models that share a tokenizer overlap. (b) Change in bits per character against the measured re-split rate; dashed: least-squares line without SmolLM3. SmolLM3 points below zero at low re-split are the renormalization effect discussed in the text.}
\label{fig:quality}
\end{figure}

Table~\ref{tab:quality} and Figure~\ref{fig:quality} show what trimming does to the models themselves.

\paragraph{The script rule.} Across the \QScriptN{} model and language pairs, bits per character change by between \QScriptBPCmin{} and \QScriptBPCmax, at least \QScriptGreedyMin{} of greedy continuations are identical to the original model's (median 100\%), and Belebele answers are identical in \QScriptBBsame{} cases and differ by one question in the rest. When the tokenization is unchanged, the trimmed model is the original model renormalized over the kept rows, so every token's probability can only rise and bits per character can only fall. The few small increases are in Korean and French, whose held-out text contains characters outside the kept scripts, such as Han characters in Korean, which the trimmed tokenizer spells in bytes (0.3\% more tokens).

\paragraph{The corpus rule and the certificate.} The certificate predicts the re-split rate on the evaluation lines as it did on the tokenizer experiments (correlation \QCertR, measured rate a median \QCertRatio{} times the prediction), and the change in bits per character follows the re-split rate almost linearly. For the five models other than SmolLM3, the least-squares line is $\Delta\mathrm{BPC} \approx \QFitSlope \times \textrm{re-split}$ \QFitIcpt (correlation \QFitR, n=\QFitN): each percent of re-split tokens costs about 2.5\% in bits per character, and renormalization gives back about 1\%. The certificate therefore predicts the quality cost before any model is run. In the \QOkN{} settings that it certifies (predicted re-split under 1\%), the corpus rule removes a median \QOkRm{} of rows, bits per character change by a median \QOkBPCmed{} and at most \QOkBPCmax, and a median \QOkGreedy{} of greedy continuations are identical. In the \QMoreN{} settings where it asks for more text, the median change is \QMoreBPCmed, and the worst, English with 4,000 lines, reaches \QMaxBPC{} at a re-split rate of \QMaxRs. Greedy agreement is the strictest of the three measures, since one different token among 24 counts as a mismatch; it falls with the re-split rate in the same way.

\paragraph{Renormalisation is a property of the model.} SmolLM3 uses the Llama~3 tokenizer, so its keep sets and re-split rates are identical to those of Llama~3.2 1B. Yet where the re-split rate is under 0.2\%, its bits per character change by \QSmolLo{} to \QSmolHi, against falls of at most 2.3\% for Llama~3.2 1B. With the tokenization unchanged, the change equals the log of the probability the original model gives to kept pieces, so SmolLM3 places far more probability on pieces that never occur in the target language. This is not a better model: its Belebele accuracy does not rise. A drop in bits per character after trimming should therefore be read together with the re-split rate, and comparisons of trimmed models are clearest when made against the same model's script rule, which isolates the renormalization.

\paragraph{Belebele.} These models are close to chance on Belebele (0.20--0.39 against 0.25), and with 200 questions two binomial standard errors are about 0.06. Only \QBBbeyond{} of \QSettings{} changes exceed that, all drops for Llama~3.2 1B under the corpus rule in Chinese and Japanese (largest \QBBmaxDrop). Belebele passages come from Wikipedia through FLORES, not from the treebanks the keep sets were built on, and we did not measure the re-split rate on them; the drop is consistent with the domain effect of Table~\ref{tab:domain}, which the 200-line check on deployment text is meant to catch. At this model scale Belebele detects only large changes; bits per character and greedy agreement are the more sensitive measures.

\subsection{Trimming moves edge models under memory budgets}
\label{sec:rq5}

\begin{table}[t]
\centering\small
\caption{Edge-sized models in P2 (n=\FlipN) that exceed a weight budget, and how many fit after trimming to Thai and English. Weight bytes at the stored precision, or recomputed at INT8 or INT4. Corpus rule: merge-closed, 15,000 Thai lines.}
\label{tab:flips}
\begin{tabular}{lrrr}
\toprule
Budget & Over budget & Fit after script rule & Fit after corpus rule \\
\midrule
Stored precision, 1\,GB & 182 & 3 & 25 \\
Stored precision, 2\,GB & 131 & 5 & 16 \\
Stored precision, 4\,GB & 64 & 9 & 19 \\
INT8, 1\,GB & 124 & 3 & 6 \\
INT8, 2\,GB & 54 & 10 & 21 \\
INT4, 1\,GB & 54 & 10 & 21 \\

\bottomrule
\end{tabular}
\end{table}

\begin{table}[t]
\centering\small
\caption{Reference models. Tax as a share of stored model bytes (GB removed in brackets) for Thai and English, for Thai, English and Chinese, and under the merge-closed Thai corpus rule (15,000 lines; families outside P2: --).}
\label{tab:models}
\begin{tabular}{lrrrr}
\toprule
Model & Size (GB) & Script TH+EN (\%) & Script TH+EN+ZH (\%) & Corpus TH (\%) \\
\midrule
Gemma 3 270M & 0.54 & 21.9 (0.12) & 17.3 & 55.4 (0.30) \\
Qwen3 0.6B & 1.50 & 11.9 (0.18) & 5.0 & 35.8 (0.54) \\
Gemma 3 1B & 2.00 & 10.6 (0.21) & 8.4 & 26.7 (0.53) \\
Qwen3 1.7B & 4.06 & 8.8 (0.36) & 3.7 & 26.5 (1.08) \\
Whisper large-v3-turbo & 1.62 & 1.2 (0.02) & 0.9 & -- \\
Gemma 4 E2B & 10.25 & 18.8 (1.93) & 14.9 & 47.5 (4.87) \\
Ministral 3 3B & 4.67 & 4.7 (0.22) & 4.2 & -- \\
Whisper medium & 3.06 & 1.0 (0.03) & 0.8 & -- \\

\bottomrule
\end{tabular}
\end{table}

Table~\ref{tab:flips} counts edge-sized models of P2 whose weights exceed a budget and fit after trimming. At stored precision and a 1\,GB budget, \FlipSAs{} of \FlipSAe{} over-budget models fit after the script rule and \FlipSAc{} after the corpus rule; at INT4 and 1\,GB, \FlipFAs{} and \FlipFAc{} of \FlipFAe. Qwen3 0.6B (1.50\,GB) fits in 1\,GB only after the corpus rule, and so does Qwen3 1.7B at INT4. Table~\ref{tab:models} shows reference models. Adding Chinese to Thai and English reduces the tax of Qwen3 0.6B from \ExQwenSmallS{} to 5.0\%, because Chinese pieces are a large share of its vocabulary, but reduces that of Gemma~3 270M only from 21.9\% to 17.3\%.

\section{Discussion and limitations}
\label{sec:discussion}

\paragraph{Model quality.} Section~\ref{sec:quality} measures quality without retraining, on models up to 3B parameters and on text from the same source as the trimming text. Larger models, generation quality judged by people, and fine-tuning after trimming are not covered, and Belebele has little power at this scale. The relation between re-split rate and bits per character was close to linear for five of six models, but its intercept depends on how much probability a model places outside the kept pieces, as SmolLM3 shows. The certificate estimates how much of the input changes at all, and merge closure matters for the same reason: without it, two to three times more of the input changes than a user who checks the nominal rate would believe.

\paragraph{Bytes on disk and bytes in memory.} The tax is measured in stored bytes. Runtimes that memory-map weights page in only the embedding rows they touch, so for them the input embedding costs disk and download size more than memory, while the output head is read in full at every decoding step and costs memory bandwidth and compute. The deployment counts of Table~\ref{tab:flips} are therefore for weight files; the reduction in resident memory depends on the runtime and on whether the head is tied.

\paragraph{Scope.} Tokenizer families are assigned by model type and vocabulary size, which can misassign a model that ships a custom tokenizer with a standard size; the package reads each model's own tokenizer when it is used directly. The certificate is for in-domain text; on a shifted domain the real rate was a median \DomRatio{} times higher in our test, and a 200-line nominal check on deployment text is the recommended safeguard. The newest Hub-only tokenizer families (Qwen3.5 and later, Nemotron-H, GLM-5, DeepSeek-V3.2 and V4) are not in the census populations. Treebanks for Tamil, Bengali, Marathi and Telugu are too small for the corpus experiments. The corpus-rule experiments cover the four BPE families we rebuild from merge lists; the package also closes tiktoken, Tekken and SentencePiece-BPE vocabularies, taking the parents of a SentencePiece-BPE piece from the highest-score merge order (on Thai text no reachable kept piece of the Mistral v3 or Llama 2 vocabulary was left unreachable after closure), and SentencePiece Unigram needs no closure. These extensions are not part of the measured experiments. Encoder models and the vision and audio towers of multimodal models are outside the populations studied here.

\section{Conclusion}

Multilingual vocabularies have a memory cost that single-language deployments pay in full. Measured from manifests and tokenizer files, it is a few percent of a large model and over a quarter of a small one under the corpus rule, it is about the same for every language, and it trades against the extra tokens that small vocabularies cost non-English languages. Removing it by writing system is exact when the keep rule includes the shared character classes and Latin. Removing it with the user's text goes further, but only if the keep set is closed under BPE merges; the usual frequency rule silently strands rows and changes two to three times more of the input than its check reports. With closure, the amount of text a deployment needs is predictable from that text itself: a Good--Turing estimate and a fitted decay tell the user, before deployment, how much of their future input the trimmed tokenizer will split differently and how much more text would fix it, and on the models we tested the loss in bits per character follows that rate. The census, the merge-closed trimming and the certificate are available in \tn{ai-model-explorer}, and the tables behind this paper in the \tn{ai-vocabulary-tax} dataset.

\bibliographystyle{plainnat}
\bibliography{refs}

\appendix
\section{Script rule for 40 languages}
\label{app:script}

\begin{table}[H]
\centering\footnotesize
\caption{Median vocabulary tax (\% of model bytes) under the script rule (target language plus Latin) in P1, by model size; download-weighted tax of edge-sized models; tax of Gemma~4 E2B in GB. Languages with identical rows are grouped.}
\label{tab:app-script}
\begin{tabularx}{\linewidth}{Xrrrrrr}
\toprule
Languages & $<$0.5B & 0.5--2B & 2--8B & $>$8B & Edge, weighted & Gemma 4 E2B (GB) \\
\midrule
Dutch, English, French, German, Indonesian, Italian, Lao, Malay, Polish, Portuguese, Punjabi, Spanish, Swahili, Tagalog, Turkish, Vietnamese & 4.5 & 4.6 & 2.9 & 2.2 & 5.1 & 1.97 \\
Bengali & 4.5 & 4.6 & 2.9 & 2.2 & 5.1 & 1.77 \\
Telugu & 4.5 & 4.6 & 2.9 & 2.2 & 5.1 & 1.94 \\
Gujarati, Kannada & 4.5 & 4.6 & 2.9 & 2.2 & 5.1 & 1.95 \\
Amharic, Burmese, Khmer & 4.5 & 4.6 & 2.9 & 2.2 & 5.1 & 1.96 \\
Thai & 4.4 & 4.3 & 2.8 & 2.1 & 4.8 & 1.93 \\
Arabic, Persian, Urdu & 4.4 & 4.2 & 2.7 & 1.8 & 4.7 & 1.80 \\
Hindi, Marathi & 4.4 & 4.6 & 2.9 & 2.1 & 5.1 & 1.69 \\
Tamil & 4.4 & 4.6 & 2.9 & 2.2 & 5.1 & 1.89 \\
Malayalam & 4.4 & 4.6 & 2.9 & 2.2 & 5.1 & 1.93 \\
Hebrew & 4.4 & 4.3 & 2.8 & 2.2 & 4.8 & 1.96 \\
Greek & 4.4 & 4.6 & 2.9 & 2.1 & 5.0 & 1.94 \\
Georgian & 4.4 & 4.6 & 2.9 & 2.2 & 5.1 & 1.96 \\
Armenian & 4.4 & 4.6 & 2.9 & 2.2 & 5.1 & 1.97 \\
Korean & 4.0 & 4.2 & 2.7 & 2.0 & 4.7 & 1.88 \\
Chinese & 3.6 & 2.5 & 1.4 & 1.5 & 2.6 & 1.57 \\
Japanese & 3.3 & 2.2 & 1.3 & 1.3 & 2.3 & 1.47 \\
Russian, Ukrainian & 2.4 & 4.2 & 2.7 & 1.5 & 4.5 & 1.69 \\

\bottomrule
\end{tabularx}
\end{table}

\section{Script rule validation}
\label{app:valid}
\begin{table}[H]
\centering\scriptsize\setlength{\tabcolsep}{2.6pt}
\caption{Vocabulary rows removed (\%) when keeping each PUD language plus Latin, for nine families. The share of held-out tokens in removed rows is at most \MissMax{} in every cell.}
\label{tab:app-valid}
\begin{tabular}{lrrrrrrrrrrrrrrrrrrrrr}
\toprule
Family & ar & cs & de & en & es & fi & fr & gl & hi & id & is & it & ja & ko & pl & pt & ru & sv & th & tr & zh \\
\midrule
Gemma 3/3n/4 & 33 & 36 & 36 & 36 & 36 & 36 & 36 & 36 & 31 & 36 & 36 & 36 & 27 & 34 & 36 & 36 & 31 & 36 & 35 & 36 & 29 \\
Qwen 2--3.5 & 28 & 30 & 30 & 30 & 30 & 30 & 30 & 30 & 30 & 30 & 30 & 30 & 12 & 28 & 30 & 30 & 28 & 30 & 29 & 30 & 14 \\
Llama 3 & 14 & 17 & 17 & 17 & 17 & 17 & 17 & 17 & 16 & 17 & 17 & 17 & 13 & 15 & 17 & 17 & 12 & 17 & 16 & 17 & 13 \\
Llama 2 & 13 & 13 & 13 & 13 & 13 & 13 & 13 & 13 & 13 & 13 & 13 & 13 & 11 & 13 & 13 & 13 & 4 & 13 & 13 & 13 & 11 \\
Mistral Tekken & 20 & 28 & 28 & 28 & 28 & 28 & 28 & 28 & 26 & 28 & 28 & 28 & 24 & 24 & 28 & 28 & 22 & 28 & 27 & 28 & 25 \\
Whisper & 14 & 14 & 14 & 14 & 14 & 14 & 14 & 14 & 14 & 14 & 14 & 14 & 11 & 11 & 14 & 14 & 9 & 14 & 14 & 14 & 12 \\
GPT-2 & 0 & 0 & 0 & 0 & 0 & 0 & 0 & 0 & 0 & 0 & 0 & 0 & 0 & 0 & 0 & 0 & 0 & 0 & 0 & 0 & 0 \\
XLM-R & 48 & 54 & 54 & 54 & 54 & 54 & 54 & 54 & 51 & 54 & 54 & 54 & 45 & 51 & 54 & 54 & 41 & 54 & 52 & 54 & 47 \\
Mistral v3 & 12 & 12 & 12 & 12 & 12 & 12 & 12 & 12 & 12 & 12 & 12 & 12 & 8 & 11 & 12 & 12 & 7 & 12 & 12 & 12 & 8 \\

\bottomrule
\end{tabular}
\end{table}

\section{Corpus rule for 22 languages}
\label{app:corpus}
\begin{table}[H]
\centering\footnotesize
\caption{Corpus rule with 4,000 trimming lines (2,000 for Urdu and Chinese): rows removed with merge closure (\%) / real re-split rate with closure (\%) / real re-split rate without closure (\%).}
\label{tab:app-corpus}
\begin{tabular}{lrrrr}
\toprule
Language & Gemma & Qwen & Llama 3 & GPT-2 \\
\midrule
Arabic & 92/0.14/0.23 & 91/0.05/1.09 & 90/0.03/0.05 & 94/0.00/0.00 \\
Chinese (2k) & 88/3.25/3.69 & 86/7.24/10.19 & 89/1.01/1.43 & 92/0.25/0.29 \\
Czech & 90/2.72/4.76 & 88/1.38/2.25 & 84/2.14/3.96 & 83/1.06/1.61 \\
Dutch & 89/4.10/7.77 & 88/2.74/4.88 & 85/2.81/4.51 & 81/1.86/2.99 \\
English (web) & 86/8.71/26.93 & 83/7.64/26.18 & 78/7.74/23.54 & 66/7.21/25.37 \\
French & 86/5.06/10.57 & 85/3.08/5.71 & 81/2.99/5.00 & 74/2.14/3.82 \\
German & 88/3.90/10.57 & 86/2.26/7.14 & 83/2.26/5.14 & 79/1.40/3.32 \\
Hebrew & 93/0.00/0.00 & 91/0.15/0.37 & 92/0.00/0.00 & 94/0.00/0.00 \\
Hindi & 90/2.01/9.64 & 93/0.00/0.00 & 92/0.00/0.01 & 94/0.00/0.00 \\
Indonesian & 87/5.05/9.92 & 86/2.77/4.54 & 82/2.72/4.46 & 76/1.98/2.95 \\
Italian & 88/3.64/8.56 & 87/2.04/5.08 & 84/1.86/3.66 & 80/1.46/2.76 \\
Japanese & 87/1.56/2.21 & 89/1.18/1.97 & 89/0.39/0.56 & 92/0.12/0.17 \\
Korean & 92/0.56/0.69 & 91/0.42/0.65 & 90/0.40/0.51 & 93/0.04/0.05 \\
Persian & 92/0.46/1.36 & 92/0.11/0.25 & 90/0.17/0.67 & 94/0.02/0.02 \\
Polish & 90/2.46/4.77 & 88/1.62/2.84 & 87/1.45/2.08 & 85/0.95/1.20 \\
Portuguese & 87/4.21/10.18 & 86/2.29/5.09 & 83/2.13/4.19 & 79/1.46/2.58 \\
Russian & 90/1.52/3.43 & 90/0.32/1.58 & 88/0.46/1.15 & 93/0.06/0.08 \\
Spanish & 86/3.58/10.16 & 86/1.58/4.65 & 83/1.54/3.74 & 79/1.04/2.10 \\
Thai & 91/1.59/3.43 & 89/0.92/1.98 & 87/1.05/1.78 & 86/0.28/0.49 \\
Turkish & 90/3.56/6.75 & 88/2.18/4.80 & 85/2.67/5.52 & 84/1.31/2.07 \\
Ukrainian & 91/1.11/1.71 & 91/0.24/0.47 & 88/0.56/1.19 & 92/0.06/0.08 \\
Urdu (2k) & 92/0.79/3.61 & 92/0.09/0.35 & 91/0.21/0.47 & 94/0.01/0.01 \\

\bottomrule
\end{tabular}
\end{table}

\section{Ablations}
\label{app:ablations}
\paragraph{Keeping Latin.} Trimming to the Thai script alone, without Latin, loses 6.6\% (Gemma), 5.0\% (Qwen), 5.7\% (Llama~3), 5.4\% (Tekken), 4.2\% (Whisper), 3.6\% (Mistral v3), 3.5\% (Llama~2) and 1.8\% (GPT-2) of the tokens in held-out Thai text; with Latin kept, none.

\paragraph{Rank prior.} Nominal re-split rates, before closure. Keeping the first $K$ pieces by vocabulary order in addition to the corpus rule, with 1,000 trimming lines: for Qwen and French the re-split rate is 8.6, 7.5, 6.6 and 5.3\% for $K=0$, 4k, 8k and 16k, with 88.9, 87.5, 85.6 and 81.3\% of rows removed. Increasing the trimming text to 4,000 lines with $K=0$ gives 3.2\% at 85.5\%. The pattern is the same for the other families and languages.

\section{Reproducibility}
The results were produced with version 1.0.7 of the package on PyPI. The package computes the tax with the \tn{vocab\_tax} method of \tn{explore(repo)}, given the languages and optionally the user's text, and trims with \tn{trim\_vocab}, which also takes an output folder; both close the keep set under BPE merges (\tn{merge\_closure\_added}) and return the certificate fields \tn{predicted\_resplit\_rate}, \tn{lines\_needed\_for\_target} and \tn{certificate}. The browser tool (Where to cut, on the playground) encodes text with \tn{tokenizer.json} (BPE and Unigram), tiktoken, Tekken, SentencePiece and GGUF vocabularies; against \tn{tokenizers}, \tn{tiktoken}, \tn{sentencepiece} and llama.cpp it matches token for token on all 605 test sentences for 19 of the 24 vocabularies we checked and on at least 603 for four more (SentencePiece normalization rules), with DeepSeek-LLM's pre-tokenizer the exception (564), and it computes the same keep set and certificate as the package. Model quality is measured with \tn{quality\_eval.py}, released with the experiment scripts. The tables are published at \url{\DataURL}: \tn{models} and \tn{model\_language} (tax per model, language and rule), \tn{tokenizers} (rows removed and real re-split rate per family, language and amount of text), \tn{certificate}, \tn{merge\_closure} and \tn{quality}. The command line equivalent is
\begin{center}\tn{ai-model-explorer tax REPO --lang th --corpus mine.txt --target 0.01 --export out/}\end{center} Treebank text is used under the licenses of the individual treebanks.

\end{document}

%% file: numbers.tex
\newcommand{\NMapped}{2{,}171}
\newcommand{\NGen}{1{,}244}

\newcommand{\TaxSA}{4.4\%}
\newcommand{\TaxSnA}{116}
\newcommand{\TaxSciA}{[2.3, 7.9]}
\newcommand{\TaxSpnA}{22.1\%}
\newcommand{\TaxSB}{4.3\%}

\newcommand{\TaxSC}{2.8\%}

\newcommand{\TaxSD}{2.1\%}
\newcommand{\TaxSnD}{628}

\newcommand{\TaxDLW}{4.8\%}
\newcommand{\NCorpusPop}{909}

\newcommand{\TaxCA}{29.0\%}
\newcommand{\TaxCsA}{7.9\%}

\newcommand{\TaxCD}{9.1\%}
\newcommand{\TaxCsD}{2.1\%}

\newcommand{\ExGemmaFourS}{18.8\%}
\newcommand{\ExGemmaFourSGB}{1.93}
\newcommand{\ExGemmaFourC}{47.5\%}
\newcommand{\ExGemmaFourCGB}{4.87}
\newcommand{\ExGemmaFourGB}{10.25}

\newcommand{\ExQwenSmallS}{11.9\%}

\newcommand{\ExQwenSmallC}{35.8\%}

\newcommand{\ThRemLo}{79\%}
\newcommand{\ThRemHi}{89\%}
\newcommand{\ThRsLo}{0.10\%}
\newcommand{\ThRsHi}{0.67\%}
\newcommand{\FlipN}{257}
\newcommand{\FlipSAe}{182}
\newcommand{\FlipSAs}{3}
\newcommand{\FlipSAc}{25}

\newcommand{\FlipFAe}{54}
\newcommand{\FlipFAs}{10}
\newcommand{\FlipFAc}{21}

\newcommand{\NaiveRatio}{1.9}
\newcommand{\ClosedRatio}{0.94}
\newcommand{\NaiveTok}{2.5\%}
\newcommand{\ClosedTok}{0.9\%}
\newcommand{\NaiveTokMax}{62\%}

\newcommand{\ExtraRows}{5.9\%}
\newcommand{\ExtraRowsMax}{36\%}
\newcommand{\OrphMed}{828}
\newcommand{\OrphMin}{13}
\newcommand{\OrphMax}{6{,}198}
\newcommand{\OrphShare}{5.0\%}
\newcommand{\NaiveWorse}{95}
\newcommand{\NaiveMeas}{101}
\newcommand{\CertN}{300}
\newcommand{\CertR}{0.996}
\newcommand{\CertRatio}{1.01}
\newcommand{\CertIQR}{0.94--1.06}
\newcommand{\CertAgree}{297}
\newcommand{\CertGTA}{6.6\%}
\newcommand{\CertSpA}{22.6\%}

\newcommand{\CertGTB}{5.9\%}
\newcommand{\CertSpB}{24.4\%}

\newcommand{\CertGTC}{4.5\%}
\newcommand{\CertSpC}{25.9\%}

\newcommand{\ExtN}{35}
\newcommand{\ExtRatio}{1.03}
\newcommand{\ExtIQR}{0.94--1.12}
\newcommand{\ExtRatioHi}{1.06}
\newcommand{\ExtNHi}{30}

\newcommand{\MLCorpLo}{27.8\%}
\newcommand{\MLCorpHi}{31.1\%}

\newcommand{\RSAr}{0.04\%}
\newcommand{\RSRu}{0.39\%}
\newcommand{\RSTh}{0.99\%}
\newcommand{\RSDe}{2.26\%}
\newcommand{\RSFr}{3.04\%}
\newcommand{\RSEn}{7.69\%}

\newcommand{\RPa}{8.6\%}
\newcommand{\RPb}{5.3\%}
\newcommand{\RPcost}{7.6}
\newcommand{\MissMax}{0.008\%}

%% file: numbers_base.tex
\newcommand{\MatchVT}{1.30}
\newcommand{\MatchFreq}{1.32}
\newcommand{\MatchRank}{8.4}
\newcommand{\MatchShareVT}{91\%}
\newcommand{\MatchShareFreq}{90\%}
\newcommand{\OrcOurs}{52.6\%}
\newcommand{\OrcVT}{46.6\%}
\newcommand{\OrcFreq}{46.5\%}
\newcommand{\OrcRank}{22.4\%}
\newcommand{\OrcScript}{17.3\%}
\newcommand{\ViolVTtwo}{13}
\newcommand{\ViolFreqtwo}{9}
\newcommand{\ViolOurstwo}{0}
\newcommand{\ViolVTone}{6}
\newcommand{\ViolFreqone}{3}
\newcommand{\ViolOursone}{3}
\newcommand{\CVerr}{25--27\%}
\newcommand{\ExtFitErr}{0.10}
\newcommand{\ExtBoneErr}{0.32}

%% file: numbers_extra.tex
\newcommand{\DomRatio}{1.7}
\newcommand{\DomRatioMax}{3.4}
\newcommand{\DomNom}{0.86}
\newcommand{\DomAdd}{6\%}
\newcommand{\MixPer}{1.03}
\newcommand{\MixPerIQR}{0.96--1.10}
\newcommand{\MixPerN}{40}
\newcommand{\MixPool}{1.02}
\newcommand{\MixNaive}{1.65}
\newcommand{\WasteScript}{2.8\%}
\newcommand{\WasteScriptLo}{1.5\%}
\newcommand{\WasteCorpus}{8.2\%}
\newcommand{\WasteDl}{318}

%% file: numbers_quality.tex
\newcommand{\QSettings}{188}
\newcommand{\QCorpN}{122}
\newcommand{\QScriptN}{66}
\newcommand{\QScriptBPCmin}{$-$0.71\%}
\newcommand{\QScriptBPCmax}{$+$0.55\%}
\newcommand{\QScriptGreedyMin}{93\%}
\newcommand{\QScriptBBsame}{61}

\newcommand{\QCertR}{0.994}
\newcommand{\QCertRatio}{0.88}
\newcommand{\QOkN}{68}

\newcommand{\QOkBPCmed}{$-$0.65\%}
\newcommand{\QOkBPCmax}{$+$1.99\%}
\newcommand{\QOkGreedy}{91\%}
\newcommand{\QOkRm}{89.7\%}
\newcommand{\QMoreN}{54}
\newcommand{\QMoreBPCmed}{$+$3.41\%}

\newcommand{\QFitR}{0.975}
\newcommand{\QFitSlope}{2.5}
\newcommand{\QFitIcpt}{$-$1.1\%}
\newcommand{\QFitN}{101}
\newcommand{\QSmolLo}{$-$8.0\%}
\newcommand{\QSmolHi}{$-$13.8\%}
\newcommand{\QBBmaxDrop}{0.075}
\newcommand{\QBBbeyond}{3}

\newcommand{\QMaxRs}{8.8\%}
\newcommand{\QMaxBPC}{$+$23.3\%}

%% file: refs.bib
@misc{singkul2026manifest,
  title={Reading the Architecture off the Weights: Manifest-Only Model Anatomy and a Census of 12,974 {Hugging Face} Repositories},
  author={Singkul, Sattaya},
  year={2026},
  note={Preprint, ResearchGate},
  doi={10.13140/RG.2.2.19473.19048},
  howpublished={\url{https://doi.org/10.13140/RG.2.2.19473.19048}}
}

@inproceedings{ushio2023vt,
  title={Efficient Multilingual Language Model Compression through Vocabulary Trimming},
  author={Ushio, Asahi and Zhou, Yi and Camacho-Collados, Jose},
  booktitle={Findings of the Association for Computational Linguistics: EMNLP 2023},
  year={2023}
}

@inproceedings{bogoychev2024updowns,
  title={The Ups and Downs of Large Language Model Inference with Vocabulary Trimming by Language and by Frequency},
  author={Bogoychev, Nikolay and Chen, Pinzhen and Haddow, Barry and Birch, Alexandra},
  booktitle={Proceedings of the Fifth Workshop on Insights from Negative Results in NLP},
  year={2024}
}

@misc{squeezebits2025trim,
  title={Vocabulary Trimming: An Easy and Effective Method for {SLM} Acceleration},
  author={{SqueezeBits}},
  howpublished={\url{https://blog.squeezebits.com/vocabulary-trimming-methods}},
  year={2025}
}

@article{zhang2025vocabtailor,
  title={{VocabTailor}: Dynamic Vocabulary Selection for Downstream Tasks in Small Language Models},
  author={Zhang, Hanling and Zhou, Yayu and Fang, Tongcheng and Yuan, Zhihang and Dai, Guohao and Wang, Yu},
  journal={arXiv preprint arXiv:2508.15229},
  year={2025}
}

@article{kwek2025compact,
  title={{COMPACT}: Common-Token Optimized Model Pruning Across Channels and Tokens},
  author={Kwek, Eugene and Yin, Wenpeng},
  journal={arXiv preprint arXiv:2509.06836},
  year={2025}
}

@article{gurgurov2025compression,
  title={On Multilingual Encoder Language Model Compression for Low-Resource Languages},
  author={Gurgurov, Daniil and Gregor, Michal and van Genabith, Josef and Ostermann, Simon},
  journal={arXiv preprint arXiv:2505.16956},
  year={2025}
}

@article{benshoham2026spec,
  title={Balancing Coverage and Draft Latency in Vocabulary Trimming for Faster Speculative Decoding},
  author={Ben Shoham, Ofir},
  journal={arXiv preprint arXiv:2603.05210},
  year={2026}
}

@inproceedings{petrov2023unfairness,
  title={Language Model Tokenizers Introduce Unfairness Between Languages},
  author={Petrov, Aleksandar and La Malfa, Emanuele and Torr, Philip H. S. and Bibi, Adel},
  booktitle={Advances in Neural Information Processing Systems},
  year={2023}
}

@inproceedings{ahia2023cost,
  title={Do All Languages Cost the Same? Tokenization in the Era of Commercial Language Models},
  author={Ahia, Orevaoghene and Kumar, Sachin and Gonen, Hila and Kasai, Jungo and Mortensen, David and Smith, Noah and Tsvetkov, Yulia},
  booktitle={Proceedings of the 2023 Conference on Empirical Methods in Natural Language Processing},
  year={2023}
}

@inproceedings{rust2021tokenizer,
  title={How Good is Your Tokenizer? On the Monolingual Performance of Multilingual Language Models},
  author={Rust, Phillip and Pfeiffer, Jonas and Vuli{\'c}, Ivan and Ruder, Sebastian and Gurevych, Iryna},
  booktitle={Proceedings of the 59th Annual Meeting of the Association for Computational Linguistics},
  year={2021}
}

@inproceedings{sennrich2016bpe,
  title={Neural Machine Translation of Rare Words with Subword Units},
  author={Sennrich, Rico and Haddow, Barry and Birch, Alexandra},
  booktitle={Proceedings of the 54th Annual Meeting of the Association for Computational Linguistics},
  year={2016}
}

@inproceedings{kudo2018sentencepiece,
  title={{SentencePiece}: A simple and language independent subword tokenizer and detokenizer for Neural Text Processing},
  author={Kudo, Taku and Richardson, John},
  booktitle={Proceedings of the 2018 Conference on Empirical Methods in Natural Language Processing: System Demonstrations},
  year={2018}
}

@article{radford2019gpt2,
  title={Language Models are Unsupervised Multitask Learners},
  author={Radford, Alec and Wu, Jeffrey and Child, Rewon and Luan, David and Amodei, Dario and Sutskever, Ilya},
  journal={OpenAI technical report},
  year={2019}
}

@article{good1953,
  title={The Population Frequencies of Species and the Estimation of Population Parameters},
  author={Good, I. J.},
  journal={Biometrika},
  volume={40},
  number={3--4},
  pages={237--264},
  year={1953}
}

@article{gale1995gt,
  title={Good-{Turing} Frequency Estimation Without Tears},
  author={Gale, William A. and Sampson, Geoffrey},
  journal={Journal of Quantitative Linguistics},
  volume={2},
  number={3},
  pages={217--237},
  year={1995}
}

@book{heaps1978,
  title={Information Retrieval: Computational and Theoretical Aspects},
  author={Heaps, Harold Stanley},
  publisher={Academic Press},
  year={1978}
}

@article{demarneffe2021ud,
  title={Universal Dependencies},
  author={de Marneffe, Marie-Catherine and Manning, Christopher D. and Nivre, Joakim and Zeman, Daniel},
  journal={Computational Linguistics},
  volume={47},
  number={2},
  pages={255--308},
  year={2021}
}

@misc{wisesight,
  title={Wisesight Sentiment Corpus},
  author={Suriyawongkul, Arthit and Chuangsuwanich, Ekapol and Chormai, Pattarawat and Polpanumas, Charin},
  howpublished={\url{https://github.com/PyThaiNLP/wisesight-sentiment}},
  year={2019}
}

@misc{unicode_uax24,
  title={Unicode Standard Annex \#24: Unicode Script Property},
  author={{The Unicode Consortium}},
  howpublished={\url{https://www.unicode.org/reports/tr24/}}
}

@inproceedings{bandarkar2024belebele,
  title={The {Belebele} Benchmark: a Parallel Reading Comprehension Dataset in 122 Language Variants},
  author={Bandarkar, Lucas and Liang, Davis and Muller, Benjamin and Artetxe, Mikel and Shukla, Satya Narayan and Husa, Donald and Goyal, Naman and Krishnan, Abhinandan and Zettlemoyer, Luke and Khabsa, Madian},
  booktitle={Proceedings of the 62nd Annual Meeting of the Association for Computational Linguistics},
  year={2024}
}

@misc{llamacpp,
  title={llama.cpp: {LLM} inference in {C/C++}},
  author={Gerganov, Georgi and {ggml-org contributors}},
  howpublished={\url{https://github.com/ggml-org/llama.cpp}},
  note={Vocabulary files in models/, read November 2026},
  year={2023}
}

@misc{safetensors,
  title={safetensors: Simple, safe way to store and distribute tensors},
  author={{Hugging Face}},
  howpublished={\url{https://github.com/huggingface/safetensors}},
  year={2022}
}

@inproceedings{radford2023whisper,
  title={Robust Speech Recognition via Large-Scale Weak Supervision},
  author={Radford, Alec and Kim, Jong Wook and Xu, Tao and Brockman, Greg and McLeavey, Christine and Sutskever, Ilya},
  booktitle={International Conference on Machine Learning},
  year={2023}
}

@book{efron1993,
  title={An Introduction to the Bootstrap},
  author={Efron, Bradley and Tibshirani, Robert J.},
  publisher={Chapman and Hall},
  year={1993}
}

@inproceedings{land2024magikarp,
  title={Fishing for {Magikarp}: Automatically Detecting Under-trained Tokens in Large Language Models},
  author={Land, Sander and Bartolo, Max},
  booktitle={Proceedings of the 2024 Conference on Empirical Methods in Natural Language Processing},
  year={2024}
}
